%% file: VPTTA_Arxiv.tex
\documentclass[10pt,twocolumn,letterpaper]{article}

\usepackage[pagenumbers]{cvpr}   
\usepackage{float}
\usepackage[ruled,vlined]{algorithm2e}
\usepackage{algorithmic}
\usepackage{threeparttable}
\usepackage{multicol}
\usepackage{multirow}
\usepackage{booktabs} 
\usepackage{makecell} 
\usepackage{amsmath}
\usepackage{bm}
\usepackage{bbding}
\input{preamble}

\definecolor{cvprblue}{rgb}{0.21,0.49,0.74}
\usepackage[pagebackref,breaklinks,colorlinks,citecolor=cvprblue]{hyperref}

\title{Each Test Image Deserves A Specific Prompt: Continual Test-Time Adaptation for 2D Medical Image Segmentation}

\author{
Ziyang Chen$^{1}$~~~Yongsheng Pan$^{1\dag}$~~~Yiwen Ye$^{1}$~~~Mengkang Lu$^{1}$~~~Yong Xia$^{1, 2, 3\dag}$\\
$^{1}$ School of Computer Science and Engineering, Northwestern Polytechnical University, China\\
$^{2}$ Research \& Development Institute of Northwestern Polytechnical University in Shenzhen, China\\
$^{3}$ Ningbo Institute of Northwestern Polytechnical University, China\\
{\tt\small zychen@mail.nwpu.edu.cn, yspan@nwpu.edu.cn, ywye@mail.nwpu.edu.cn} \\ 
{\tt\small lmk@mail.nwpu.edu.cn, yxia@nwpu.edu.cn}
}
\begin{document}

\twocolumn[{%
\renewcommand\twocolumn[1][]{#1}%
\maketitle
\begin{center}
\centering
\includegraphics[width=\textwidth]{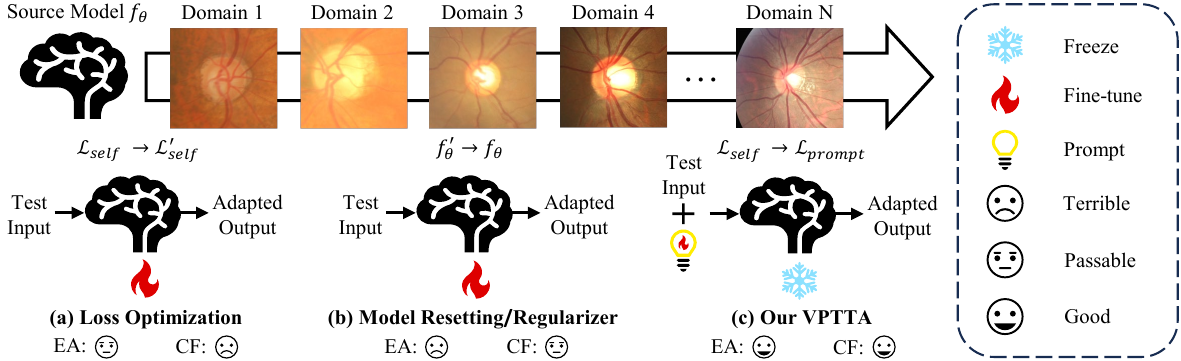}
\captionof{figure}{Comparison between our VPTTA and existing solutions under the CTTA setup. Our VPTTA avoids both error accumulation (EA) and catastrophic forgetting (CF) by freezing model parameters and achieves adaptation by training a prompt for each test image. The commonly used self-supervised loss and the optimized one are denoted by $\mathcal{L}_{self}$ and $\mathcal{L}_{self}^{'}$, respectively. 
}
\label{fig1}
\end{center}%
}]

\begin{abstract}
Distribution shift widely exists in medical images acquired from different medical centres and poses a significant obstacle to deploying the pre-trained semantic segmentation model in real-world applications.
Test-time adaptation has proven its effectiveness in tackling the cross-domain distribution shift during inference. However, most existing methods achieve adaptation by updating the pre-trained models, rendering them susceptible to error accumulation and catastrophic forgetting when encountering a series of distribution shifts (\emph{i.e.}, under the continual test-time adaptation setup).
To overcome these challenges caused by updating the models, in this paper, we freeze the pre-trained model and propose the \textbf{V}isual \textbf{P}rompt-based \textbf{T}est-\textbf{T}ime \textbf{A}daptation (VPTTA) method to train a specific prompt for each test image to align the statistics in the batch normalization layers.
Specifically, we present the low-frequency prompt, which is lightweight with only a few parameters and can be effectively trained in a single iteration.
To enhance prompt initialization, we equip VPTTA with a memory bank to benefit the current prompt from previous ones.
Additionally, we design a warm-up mechanism, which mixes source and target statistics to construct warm-up statistics, thereby facilitating the training process.
Extensive experiments demonstrate the superiority of our VPTTA over other state-of-the-art methods on two medical image segmentation benchmark tasks.
The code and weights of pre-trained source models are available at \href{https://github.com/Chen-Ziyang/VPTTA}{https://github.com/Chen-Ziyang/VPTTA}.
\end{abstract}

\renewcommand{\thefootnote}{}
\footnote{$^\dag$Yong Xia and Yongsheng Pan are the corresponding authors.
This work was supported in part by the National Natural Science Foundation of China under Grants 62171377, in part by STI2030-Major Project under Grants 2022ZD0213100, in part by Shenzhen Science and Technology Program under Grants JCYJ20220530161616036, and in part by the Ningbo Clinical Research Center for Medical Imaging under Grant 2021L003 (Open Project 2022LYKFZD06). 
}

\section{Introduction}
\label{sec:intro}
Semantic segmentation serves a pivotal role in medical image processing to outline specific anatomical structures. 
However, the segmentation model pre-trained on the source dataset may be hindered from deploying in real-world applications due to the commonly existing distribution shift, which is typically caused by the variations in imaging protocols, operators, and scanners~\cite{domain_shift_cvpr,domain_shift_2}. 
Test-time adaptation (TTA) has emerged as a promising paradigm to replace domain adaptation (DA)~\cite{FDA,MSDA,BEAL,DALN}, as TTA requires only test data during the inference phase, while DA necessitates a sufficiently large and representative target dataset~\cite{DLTTA}.
The mainstream TTA methods~\cite{TENT,TTT,MEMO,DomainAdaptor,EATA,SAR,TIPI,TeST} leverage the self-supervised tasks to construct auxiliary losses, such as entropy minimization~\cite{TENT} and self-training~\cite{TeST}, to update the parameters of pre-trained models.
Although these TTA methods successfully alleviate the domain gap and performance degradation, they are based on the assumption that the target domains exhibit static distributions.
Unfortunately, the target domains are continuously changing rather than static in most real-world scenarios. 

This was first presented as the continual test-time adaptation (CTTA) setup~\cite{CoTTA} that addresses TTA on a series of distribution shifts. 
Under this setup, pre-trained models are required to adapt to several distinct domains and therefore are more prone to suffer error accumulation and catastrophic forgetting. 
As mentioned above, many TTA methods rely on self-supervised tasks and may be affected by the noisy losses due to unreliable supervision, leading to error accumulation~\cite{Pseudo1,Pseudo2}. 
Catastrophic forgetting typically occurs when the model is continually trained on new domains, resulting in unexpected performance degradation~\cite{Forgetting,EATA}.
Some research has been studied in addressing these two challenges based on optimizing the loss functions for better supervision~\cite{RMT,CoTTA,SAR,CoTTA,DLTTA} (see Figure~\ref{fig1} (a)) or prevent the model parameters from excessive alterations by model resetting or introducing regularizers~\cite{CoTTA,SAR,EATA,EcoTTA} (see Figure~\ref{fig1} (b)).   
Despite their efforts, loss optimization cannot ensure a robust training process to avoid catastrophic forgetting during the long-term inference, insufficient or excessive updating of the model parameters may lead to inadequate adaptation or error accumulation.
Since the pre-trained model is updated by the self-supervised loss where the noise cannot be completely removed, these methods may still suffer from these two issues.

Considering the above limitations in CTTA, updating the pre-trained model during inference seems to be inappropriate. To this end, we propose the \textbf{V}isual \textbf{P}rompt-based \textbf{T}est-\textbf{T}ime \textbf{A}daptation (VPTTA) method to avoid error accumulation and catastrophic forgetting associated with updating model by freezing the model parameters and conduct adaptation by learning a specific visual prompt for each test image (see Figure~\ref{fig1} (c)). 
The prompt can adapt each test image to the frozen pre-trained model to take into account both adaptation ability and knowledge retaining.
We attempt to achieve this by answering the following three questions.

\begin{itemize}
\item How to design the prompt? Each prompt is restricted to a limited number of training iterations (sometimes just one) to satisfy the demands of online inference, requiring the prompt to be lightweight. Based on this principle, we introduce the prompt for only the low-frequency components of each sample, resulting in the low-frequency prompt. Since the low-frequency components are strongly associated with style textures and can serve as the primary source of distribution shift~\cite{FDA}, our low-frequency prompt allows for maintaining lightweight and being able to reduce distribution shifts by modifying the low-frequency components.
\item How to initialize the prompt? It is well known that model training can benefit from proper parameter initialization, resulting in not only fewer training iterations but also improved performance~\cite{Initialize1,Initialize2}. We argue that this is also applicable to prompt training and therefore construct a memory bank to store the prompts and low-frequency components of previous test images. By comparing the similarity between the low-frequency components of the current and previous images, we select several historical prompts of the most similar images to initialize the current prompt.
\item How to train the prompt? 
Since the statistics mismatch between the source and target domains in the batch normalization (BN) layers is a significant cause of distribution shift~\cite{Dynamically,BNMismatch,IBN}, we optimize the prompt with the objective of minimizing the distances between BN statistics and the statistics extracted from the features of test data for statistics alignment~\cite{ProSFDA}. 
To further address the training difficulty posed by the large distribution shift at the beginning of the inference phase, we devise a statistics-fusion-based warm-up mechanism to assist in learning the prompt from easy to hard.
\end{itemize}

To summarise, our contributions are three-fold:
    (1) We present the low-frequency prompt to adjust the test image to adapt the model instead of updating the model and therefore prevent the model from error accumulation and catastrophic forgetting.
    (2) We propose the novel VPTTA framework to alleviate the distribution shift by training a sample-specific visual prompt, which is lightweight, well-initialized, and follows a gradual learning process.
    (3) The powerful framework we provide can be effectively deployed, requiring only a single iteration and a few dozen parameters for adaptation. 
        

\section{Related Work}
\label{sec:work}

\subsection{Test-time Adaptation (TTA)}
TTA aims to adapt the model pre-trained on the source domain to the test data during inference in a source-free and online manner~\cite{TTT,TENT}. 
The mainstream TTA methods are based on model updating, which constructs self-supervised auxiliary tasks to guide the model to learn on the test data~\cite{TENT,TTT,MEMO,Transformation,Contrastive,TeST,TIPI,TTAMAE,TestFit}. 
Sun \emph{et al.}~\cite{TTT} utilized an auxiliary task to predict the rotation to assist the model learning on the test data.
Wang \emph{et al.}~\cite{TENT} presented a test-time entropy minimization scheme to reduce generalization error by reducing the entropy of model predictions on test data.
Zhang \emph{et al.}~\cite{TestFit} introduced a supplementary network to fit the test data, which can work in a plug-and-play fashion, necessitating minimal hyperparameter tuning.
Some other methods are based on BN statistics, which alleviate the distribution shifts by modifying the statistics within BN layers~\cite{Dynamically,BN,DUA}. 
Mirza \emph{et al.}~\cite{DUA} constantly adapted the statistics of the BN layers to modify the feature representations of the model.
These TTA methods improve the adaptation performance on the static target domain but neglect the continually changing target domains in most real-world scenarios.

Recently, the continual test-time adaptation (CTTA)~\cite{CoTTA} is proposed to tackle TTA under a series of continually changing target domains. Due to long-term inference, both error accumulation~\cite{Pseudo1,Pseudo2} and catastrophic forgetting~\cite{EATA,Forgetting} are more prone to occur in this setup.
Many methods~\cite{CoTTA,EcoTTA,SAR,RMT,EATA,DLTTA,DomainAdaptor} optimize the self-supervised losses or prevent excessive alterations to the model parameters via model resetting or regularizer to improve this situation but are still impacted by these issues.
Wang \emph{et al.}~\cite{CoTTA} refined pseudo labels by the weight- and augmentation-averaged predictions and randomly restored a small part of model parameters.
Niu \emph{et al.}~\cite{SAR} removed partial noisy samples with large gradients and encouraged model weights toward a flat minimum.
Yang \emph{et al.}~\cite{DLTTA} designed a dynamic strategy to adjust the learning rate for each test image.
Zhang \emph{et al.}~\cite{DomainAdaptor} presented to adaptively fuse source and test statistics and improve the entropy minimization loss to a generalized one.
In contrast, our proposed VPTTA employs a learnable prompt for each test image to mitigate distribution shifts during inference, requiring no changes to the pre-trained model and therefore alleviating the error accumulation and catastrophic forgetting.

\begin{figure*}[!htb]
   \centering
   \includegraphics[width=0.9\textwidth]{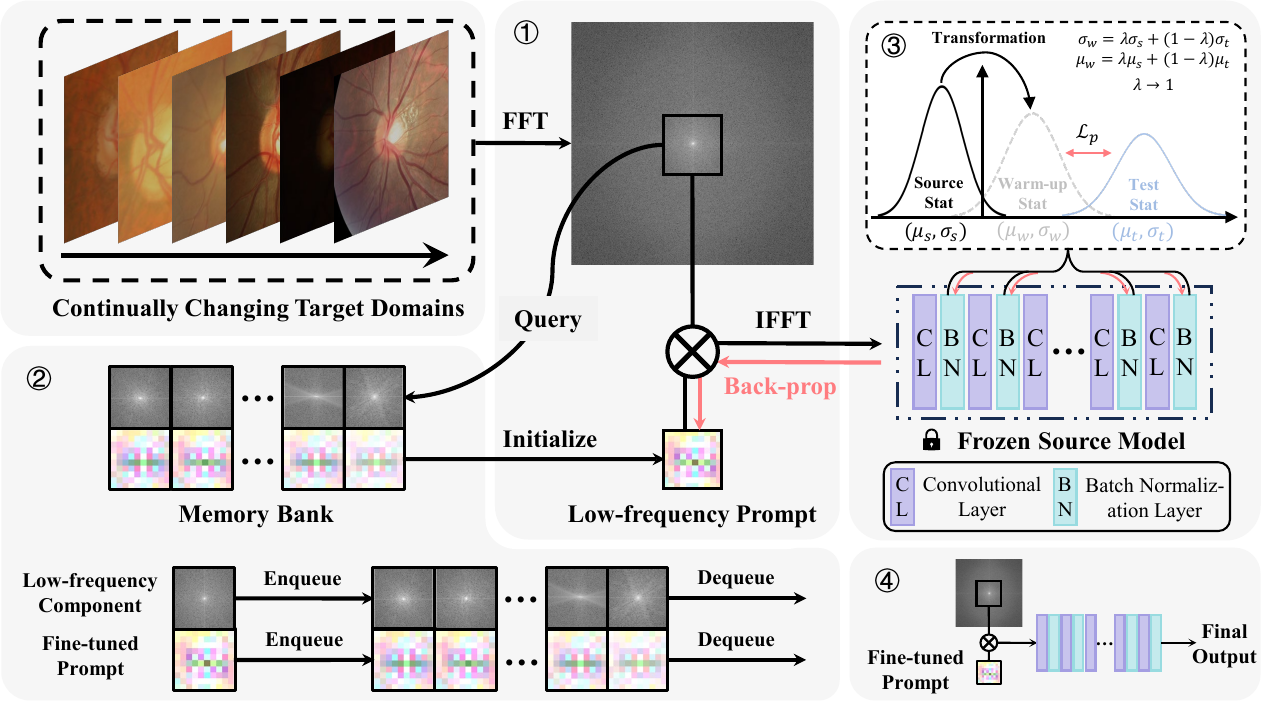}
   \caption{Overview of our VPTTA. 
   For each test image, (1) the Fast Fourier Transform (FFT) is first applied to transform it into the frequency domain, where the low-frequency component of amplitude is used to query in a memory bank to initialize the current prompt,
   and then the amplitude is multiplied with the prompt at the low-frequency component and transformed back to the spatial domain using the Inverse Fast Fourier Transform (IFFT).
   (2) The memory bank is built on the previous low-frequency components and their corresponding prompts and is updated using the First In First Out (FIFO) strategy. 
   (3) We convert the source statistics stored in BN layers into the warm-up statistics and calculate the absolute distance between warm-up and target statistics as the loss to fine-tune the prompt. 
   (4) Finally, we feed the image corrected by its fine-tuned prompt to the pre-trained model to produce the output.
   `Stat': Abbreviation of `Statistics'.
   }
   \label{fig2}
\end{figure*}

\subsection{Prompt Learning}
Prompt learning was initially employed to devise additional text instructions for input text to fine-tune the large-scale Natural Language Processing (NLP) models on downstream tasks~\cite{NLP}. With remarkable success in NLP, it has also been involved in computer vision and greatly benefited visual models in recent years~\cite{Prompt,CoCoOp,ProSFDA,DPT}.
Jia \emph{et al.}~\cite{Prompt} introduced a small number of trainable parameters in the input space while keeping the model backbone frozen to fine-tune the model with high quality and efficiency.
Zhou \emph{et al.}~\cite{CoCoOp} presented the conditional context optimization to generate an input-conditional token for each image to generalize the pre-trained vision-language model to unseen classes.
Hu \emph{et al.}~\cite{ProSFDA} employed the visual prompt at the pixel level and added it to the input images to explicitly minimize the domain discrepancy.
Gan \emph{et al.}~\cite{DPT} learned a domain-specific and domain-agnostic visual prompt for adaptation by using the exponential-moving-average strategy under the guidance of self-training.
Different from them, our VPTTA presents the low-frequency prompt to address a series of changing distribution shifts with a few dozen parameters. We further consider optimizing the initialization and warm-up mechanism, enabling our visual prompts to be effectively trained within a single iteration to satisfy the demands of CTTA.

\section{Methodology}
\label{sec:method}
\subsection{Problem Definition and Method Overview}
Let the labeled source domain dataset and unlabeled testing target dataset be 
$\mathbb{D}^{s} = \left \{ X_{i}^{s}, Y_{i}^{s} \right \}_{i=1}^{N^s}$ and $\mathbb{D}^{t} = \left \{ X_{i}^{t}\right \}_{i=1}^{N^t}$, respectively, where $X_i^{*}\in \mathbb{R}^{H \times W \times C}$ is the $i$-th image with C channels and size of $H\times W$, and $Y_i^{*}$ is its label. 
Our goal is to learn a low-frequency prompt $\mathcal{P}_i$ for each test image $X_{i}^{t}$ to obtain the modified image $\tilde{X}_{i}^{t}$ that can be adapted to the model $f_\theta:X \rightarrow Y$ pre-trained on $\mathbb{D}^{s}$.

To accomplish this goal, our VPTTA considers three key facets: prompt design, prompt initialization, and prompt training. For prompt design, the prompt is designed to change the low-frequency component of the amplitude of the test image. For prompt initialization, we adopt a memory bank to initialize the current prompt with previous ones. For prompt training, we employ the absolute distance to train the prompt to align the statistics between features extracted from test images and the batch normalization (BN) layers and also present a statistics-fusion-based warm-up mechanism.
An overview of our VPTTA is illustrated in Figure~\ref{fig2}. We now delve into its details.

\subsection{Prompt Design: Low-frequency Prompt}
Let $\mathcal{F}(\cdot)$ and $\mathcal{F}^{-1}(\cdot)$ be the Fast Fourier Transform (FFT) and Inverse Fast Fourier Transform (IFFT) operations~\cite{FFTW}, respectively. The amplitude component and phase component are denoted by $\mathcal{F}^{A}(\cdot)$ and $\mathcal{F}^{P}(\cdot)$ respectively. Note that the low-frequency component in $\mathcal{F}^{A}(\cdot)$ is shifted to the center. Utilizing the prompt $\mathcal{P}_i \in \mathbb{R}^{(\alpha \times H) \times (\alpha \times W) \times C}$ to obtain the adapted image $\tilde{X}_{i}^{t}$ can be formalized as
\begin{equation}
\begin{aligned}
    \tilde{X}_{i}^{t}=\mathcal{F}^{-1}([OnePad(\mathcal{P}_i) \odot \mathcal{F}^{A}\left ( X_i^t\right ), \mathcal{F}^{P}(X_i^t)]),
\end{aligned}
\label{eq1}
\end{equation}
where $\odot$ is the element-wise multiplication, and $OnePad$ means padding one around $\mathcal{P}_i$ to expand it to the size of $H \times W$.
Under the control of the small coefficient $\alpha \in (0,1)$, we can ensure that the prompt $\mathcal{P}_i$ remains lightweight and focuses on the low-frequency component.

Simultaneously, we recognize that the low-rank decomposition is widely employed in the field of data compression~\cite{Compression} and model fine-tuning~\cite{LoRa}. Consequently, we also design a low-rank prompt $\mathcal{P}_i^{LoRa} = \mathcal{B}_i @ \mathcal{A}_i$ as a variant (denoted as VPTTA-LR), where $\mathcal{B}_i \in \mathbb{R}^{H \times r \times C} (r\ll W)$, $\mathcal{A}_i \in \mathbb{R}^{r \times W \times C} (r\ll H)$, and $@$ denotes the matrix multiplication operation. In this way, the adapted image can be formalized as $ \tilde{X}_{i}^{t}=X_i^t+\mathcal{P}_i^{LoRa}$.

\subsection{Prompt Initialization: Memory Bank}
To achieve efficient training, it is critical to produce a suitable initialization for $\mathcal{P}_i$. Therefore, we present a memory-bank-based initialization strategy to benefit the current prompt from previous ones. 
Specifically, the memory bank $M$ stores $S$ pairs of keys and values $\left \{ q_s, v_s \right \}_{s=1}^S$ and holds the First In First Out (FIFO) principle. As shown in Figure~\ref{fig2}, the keys are the low-frequency components of previous test images, and the values are their corresponding prompts. 
Note that when using VPTTA-LR, the keys are the low-rank reconstruction matrix of previous test images.
To initialize the current prompt $\mathcal{P}_i$, we first extract the low-frequency component of the current test image $\mathcal{F}^{A}_{low}(X_i^t)$ and calculate the cosine similarity between $\mathcal{F}^{A}_{low}(X_i^t)$ and each key $q_s$ as follows:
\begin{equation}
Cos(\mathcal{F}^{A}_{low}(X_i^t), q_s) = \frac{\langle\mathcal{F}^{A}_{low}(X_i^t), q_s\rangle}{\|\mathcal{F}^{A}_{low}(X_i^t)\|*\|q_s\|},
\label{eq2}
\end{equation}
where $\langle\cdot,\cdot\rangle$ denotes the dot product operation, and $\| \cdot \|$ denotes the Euclidean norm.
Since the low-frequency component can represent the style texture of an image, we rank the similarity scores and then retrieve the values of the $K$ most relevant keys from $M$ to construct a support set $R_i=\left \{ q_k, v_k \right \}_{k=1}^K$ for the current test image. Finally, we assign a weight $w_k$ to $v_k$ to initialize the current prompt by $\mathcal{P}_i = \sum_{k=1}^{K} w_k v_k$, where the weights $\{ w_k\}$ are calculated based on the similarity scores~\cite{DLTTA} and $\sum_{k=1}^{K} w_k=1$.

\SetKwInOut{Require}{Require}
\SetKwInOut{Initialize}{Initialize}
\begin{algorithm}[t]
\caption{VPTTA algorithm.}
\small
\Initialize{Frozen source model $f_{\theta}$, initial prompt $\mathcal{P}$ filled with ones, memory bank $M$, hyperparameters $S$, $K$ and $\tau$.}
\KwIn{For each time step $i$, current test image ${X}_{i}^{t}$.}
\begin{algorithmic}[1]

\STATE $\triangleright$ \textbf{Initialize the prompt $\mathcal{P}_i$:}
\IF{$i < K$} 
\STATE $\mathcal{P}_i = \mathcal{P}$
\ELSE
\STATE Construct the support set $R_i$ from $M$
\STATE $\mathcal{P}_i = \sum_{k=1}^{K} w_k v_k, v_k \in R_i$
\ENDIF

\STATE $\triangleright$ \textbf{Train the prompt $\mathcal{P}_i$:}
\STATE Obtain $\tilde{X}_{i}^{t}$ using $\mathcal{P}_i$ by Eq.~(\ref{eq1})
\STATE Forward $f_\theta(\tilde{X}_{i}^{t})$ and calculate the warm-up statistics $\mu_{w}$ and $\sigma_{w}$ by Eq.~(\ref{eq_mu_sig}) to replace the $\mu_{s}$ and $\sigma_{s}$ stored in $f_\theta$
\STATE Backward and Update $\mathcal{P}_i$ by Eq.~(\ref{eq5})

\STATE $\triangleright$ \textbf{Inference:}
\STATE Obtain $\tilde{X}_{i}^{t}$ using updated $\mathcal{P}_i$ by Eq.~(\ref{eq1})
\STATE Forward $O_i = f_\theta(\tilde{X}_{i}^{t})$

\STATE $\triangleright$ \textbf{Update the memory bank $M$:}
\STATE $M.enqueue(q_i, v_i)$, $q_i=\mathcal{F}^{A}\left ( X_i^t\right )$, $v_i=\mathcal{P}_i$
\IF{$len(M) > S$} 
\STATE $M.dequeue(q_{first}, v_{first})$
\ENDIF

\end{algorithmic}
\KwOut{Adapted prediction $O_i$}
\label{algorithm1}
\end{algorithm}

\subsection{Prompt Training: Statistics Alignment}
Inspired by~\cite{ProSFDA,DomainAdaptor,BN}, the distribution shift largely exists in distinct BN statistics calculated on the test features (\emph{i.e.}, $\mu_t$ and ${\sigma}_t$) and stored in the source model $f_\theta$ (\emph{i.e.}, $\mu_s$ and ${\sigma}_s$), where $\mu$ and $\sigma$ respectively denote the mean and standard deviation.
To train the prompts to align the statistics, we adopt the absolute-distance-based loss $\mathcal{L}_p$~\cite{ProSFDA} as follows:
\begin{equation}
\mathcal{L}_p=\frac{1}{j}\sum|\mu_s^j-\mu_t^j|+|{\sigma}_s^j-{\sigma}_t^j|,
\label{eq3}
\end{equation}
where $j$ indicates the $j$-th BN layer in $f_\theta$.

However, aligning the source and target domains within a limited number of iterations (one in this study) provides challenges due to the empty memory bank and undiminished distribution shifts at the beginning of the inference phase.
Therefore, we design a warm-up mechanism to address this issue by simulating the warm-up statistics (\emph{i.e.}, $\mu_w$ and ${\sigma}_w$) as follows:
\begin{equation}
\mu_w = \lambda \mu_t + (1 - \lambda) \mu_s, \,
\sigma_w = \lambda \sigma_t + (1 - \lambda) \sigma_s, 
\label{eq_mu_sig}
\end{equation}
where $\lambda = \frac{1}{\sqrt{i}/\tau +1}$, $i$ indicates the index of the current test image and starts from 1, and $\tau$ is a temperature coefficient to control the rate of transition from the warm-up statistics to the source statistics.
Now, the Eq.~(\ref{eq3}) can be rewritten by:
\begin{equation}
\mathcal{L}_p=\frac{1}{j}\sum|\mu_w^j-\mu_t^j|+|{\sigma}_w^j-{\sigma}_t^j|,
\label{eq5}
\end{equation}
and we utilize $\mu_w$ and ${\sigma}_w$ for normalization instead of $\mu_s$ and ${\sigma}_s$.
As the inference progresses, the warm-up statistics will gradually shift towards the source statistics.
By simulating the warm-up statistics between the source and target statistics, we can significantly reduce the training difficulty at the beginning of the inference phase, resulting in learning prompts from easy to difficult.
The overall process of VPTTA is summarized in Algorithm ~\ref{algorithm1}.

\begin{table*}[!ht]
    \caption{Performance of our VPTTA, `Source Only' baseline, and six competing methods on the OD/OC segmentation task. The best and second-best results in each column are highlighted in \textbf{bold} and \underline{underline}, respectively.
    }
    \centering
    \resizebox{0.75\textwidth}{!}{
    \begin{tabular}{c|ccccc|c}
        \Xhline{1pt}
        {\multirow{2}*{Methods}} & 
        \multicolumn{1}{c}{Domain A} & 
        \multicolumn{1}{c}{Domain B} & 
        \multicolumn{1}{c}{Domain C} & 
        \multicolumn{1}{c}{Domain D} & 
        \multicolumn{1}{c|}{Domain E} & 
        Average \\ 
        
        \Xcline{2-7}{0.4pt}
         &
        $DSC$ &
        $DSC$ &
        $DSC$ &
        $DSC$ &
        $DSC$ &
        $DSC\uparrow$ \\
        \hline

         Source Only (ResUNet-34) &
         $64.53$ & 
         $76.06$ & 
         $71.18$ & 
         $52.67$ &
         $64.87$ &
         $65.86$ \\
         \hline

         TENT-continual (ICLR 2021)~\cite{TENT} &
         $73.07$ & 
         $78.66$ & 
         $71.94$ & 
         $46.81$ &
         $70.20$ & 
         $68.13$ \\

         CoTTA (CVPR 2022)~\cite{CoTTA} &
         $\textbf{75.39}$ &
         $75.98$ &
         $69.14$ &	
         $53.99$ &
         $70.40$ &
         $68.98$ \\ 

         DLTTA (TMI 2022)~\cite{DLTTA} &
         $\underline{75.11}$ & 	
         $\underline{78.85}$ & 	
         $\underline{73.89}$ & 	
         $51.64$ & 	
         $69.71$ & 	
         $69.84$ \\ 

         DUA (CVPR 2022)~\cite{DUA} &
         $72.28$ & 
         $76.59$ & 
         $70.13$ & 
         $56.17$ & 
         $71.38$ & 
         $69.31$ \\


         SAR (ICLR 2023)~\cite{SAR} & 
         $74.55$ &
         $77.71$ &
         $70.78$ & 		
         $55.40$ &
         $\underline{71.72}$ &
         $\underline{70.03}$ \\ 

         DomainAdaptor (CVPR 2023)~\cite{DomainAdaptor} &
         $74.50$ &
         $76.39$ &
         $71.81$ &
         $\textbf{56.78}$ &
         $70.55$ & 
         $70.01$ \\
         \hline

         VPTTA (Ours) & 
         $73.91$ & 
         $\textbf{79.36}$ & 
         $\textbf{74.51}$ & 
         $\underline{56.51}$ & 
         $\textbf{75.35}$ &
         $\textbf{71.93}$ \\
        \Xhline{1pt}
    \end{tabular}
    }
    \label{tab:Comparison_OD/OC}
\end{table*}

\begin{table*}[!htb]
    \caption{Performance of our VPTTA, `Source Only' baseline, and six competing methods on the polyp segmentation task. The best and second-best results in each column are highlighted in \textbf{bold} and \underline{underline}, respectively.
    }
    \centering
    \resizebox{1\textwidth}{!}{
    \begin{tabular}{c|ccc ccc ccc ccc|ccc}
        \Xhline{1pt}
        {\multirow{2}*{Methods}} & 
        \multicolumn{3}{c}{Domain A} & 
        \multicolumn{3}{c}{Domain B} & 
        \multicolumn{3}{c}{Domain C} & 
        \multicolumn{3}{c|}{Domain D} & 
        \multicolumn{3}{c}{Average} \\ 
        \Xcline{2-16}{0.4pt}
         &
        $DSC$ & $E^{max}_\phi$ & $S_\alpha$ &
        $DSC$ & $E^{max}_\phi$ & $S_\alpha$ &
        $DSC$ & $E^{max}_\phi$ & $S_\alpha$ &
        $DSC$ & $E^{max}_\phi$ & $S_\alpha$ &
        $DSC \uparrow$ & $E^{max}_\phi \uparrow$ & $S_\alpha \uparrow$ \\
        \hline

         Source Only (PraNet) &
         $\underline{79.90}$ & $\underline{87.97}$ & $\underline{84.66}$ &
         $66.33$ & $78.51$ & $76.72$ & 	
         $73.89$ & $84.64$ & $81.28$ & 	
         $82.95$ & $90.84$ & $88.08$ & 	
         $75.77$ & $85.49$ & $82.69$\\
         \hline

         TENT-continual (ICLR 2021)~\cite{TENT} &
         $74.86$ & $84.58$ & $80.52$ & 	
         $67.51$ & $78.66$ & $78.05$ & 	
         $17.79$ & $40.04$ & $53.30$ & 	
         $73.55$ & $83.38$ & $82.41$ & 	
         $58.43$ & $71.67$ & $73.57$ \\ 

         CoTTA (CVPR 2022)~\cite{CoTTA} &
         $76.46$ & $85.37$ & $82.56$ & 	
         $66.77$ & $76.75$ & $79.17$ & 	
         $71.39$ & $83.42$ & $80.18$ & 	
         $70.71$ & $79.81$ & $82.54$ & 	
         $71.33$ & $81.34$ & $81.11$ \\

         DLTTA (TMI 2022)~\cite{DLTTA} &
         $76.27$ & $85.23$ & $82.41$ & 	
         $66.58$ & $77.00$ & $79.24$ & 	
         $63.72$ & $78.23$ & $75.56$ & 	
         $71.20$ & $81.32$ & $83.47$ & 	
         $69.44$ & $80.45$ & $80.17$ \\ 


         DUA (CVPR 2022)~\cite{DUA} &
         $78.93$ & $87.37$ & $83.96$ & 	
         $66.84$ & $78.52$ & $77.51$ & 	
         $\underline{76.53}$ & $\underline{86.45}$ & $\underline{83.05}$ & 	
         $\underline{86.24}$ & $\underline{93.23}$ & $\underline{89.82}$ & 	
         $\underline{77.13}$ & $\underline{86.39}$ & $\underline{83.58}$ \\

         SAR (ICLR 2023)~\cite{SAR} &
         $76.48$ & $85.89$ & $81.49$ & 	
         $66.45$ & $77.35$ & $78.05$ & 	
         $71.46$ & $83.23$ & $79.40$ & 	
         $70.41$ & $80.11$ & $81.07$ & 	
         $71.20$ & $81.65$ & $80.00$ \\

         DomainAdaptor (CVPR 2023)~\cite{DomainAdaptor} &
         $77.48$ & $86.31$ & $82.40$ & 	
         $\underline{70.82}$ & $\underline{81.76}$ & $\underline{80.88}$ & 	
         $71.96$ & $83.06$ & $79.97$ & 	
         $76.89$ & $85.89$ & $84.45$ & 	
         $74.29$ & $84.26$ & $81.93$ \\ 
         \hline

         VPTTA (Ours) & 
         $\textbf{81.00}$ & $\textbf{88.91}$ & $\textbf{84.91}$ & 	
         $\textbf{76.87}$ & $\textbf{87.31}$ & $\textbf{84.08}$ & 	
         $\textbf{77.58}$ & $\textbf{87.48}$ & $\textbf{83.64}$ & 	
         $\textbf{86.39}$ & $\textbf{93.47}$ & $\textbf{89.87}$ & 	
         $\textbf{80.46}$ & $\textbf{89.29}$ & $\textbf{85.62}$ \\ 
        \Xhline{1pt}
    \end{tabular}
    }
    \label{tab:Polyp}
\end{table*}

\begin{table*}[!ht]
    \caption{Results of ablation study on the OD/OC segmentation task. The best and second-best results in each column are highlighted in \textbf{bold} and \underline{underline}, respectively.
    }
    \centering
    \resizebox{0.85\textwidth}{!}{
    \begin{tabular}{ccc|ccccc|c}
        \Xhline{1pt}
        \multicolumn{3}{c|}{Methods} & 
        \multicolumn{1}{c}{Domain A} & 
        \multicolumn{1}{c}{Domain B} & 
        \multicolumn{1}{c}{Domain C} & 
        \multicolumn{1}{c}{Domain D} & 
        \multicolumn{1}{c|}{Domain E} & 
        Average \\ 
        \hline
        Low-frequency Prompt & Memory Bank & Warm-up &
        $DSC$ & 
        $DSC$ & 
        $DSC$ & 
        $DSC$ & 
        $DSC$ & 
        $DSC\uparrow$ \\
        \hline

         & & & 
         $64.53$ & 
         $76.06$ & 
         $71.18$ & 
         $52.67$ &
         $64.87$ &
         $65.86$ \\

        \checkmark & & & 
         $65.99$ &
         $76.58$ &
         $71.75$ &
         $52.86$ &
         $67.30$ &
         $66.90$ \\

        \checkmark & \checkmark & & 
         $70.70$ &
         $\textbf{79.36}$ &
         $72.82$ &
         $\underline{55.50}$ &
         $\underline{72.84}$ &
         $\underline{70.24}$ \\

        \checkmark & & \checkmark & 
         $\textbf{74.41}$ & 
         $\underline{77.47}$ &
         $\underline{74.14}$ & 
         $54.03$ &
         $67.85$ &
         $69.58$ \\

        \checkmark & \checkmark & \checkmark & 
         $\underline{73.91}$ &
         $\textbf{79.36}$ & 
         $\textbf{74.51}$ &
         $\textbf{56.51}$ &
         $\textbf{75.35}$ & 
         $\textbf{71.93}$ \\
        \Xhline{1pt}
    \end{tabular}
    }

    \label{tab:ablation}
\end{table*}

\section{Experiments}
\label{sec:exper}
We evaluate our proposed VPTTA on two 2D segmentation benchmark tasks:
the joint optic disc (OD) and cup (OC) segmentation task, and the polyp segmentation task.
\subsection{Datasets and Evaluation Metrics}


\noindent \textbf{The OD/OC segmentation dataset} comprises five public datasets collected from different medical centres, denoted as domain A (RIM-ONE-r3~\cite{RIM_ONE_r3}), B (REFUGE~\cite{REFUGE}), C (ORIGA~\cite{ORIGA}), D (REFUGE-Validation/Test~\cite{REFUGE}), and E (Drishti-GS~\cite{Drishti-GS}). There are 159, 400, 650, 800, and 101 images from these datasets. 
We cropped a region of interest (ROI) centering at OD with size of $800\times 800$ for each image following~\cite{ProSFDA}, and each ROI is further resized to $512\times 512$ and normalized by min-max normalization. 
The Dice score metric ($DSC$) is utilized for evaluation in this task.

\noindent \textbf{The polyp segmentation dataset} consists of four public datasets collected from different medical centres, denoted as domain A (BKAI-IGH-NeoPolyp~\cite{BKAI_IGH_NeoPolyp}), B (CVC-ClinicDB~\cite{CVC_ClinicDB}), C (ETIS-LaribPolypDB~\cite{ETIS_LaribPolypDB}), and D (Kvasir-Seg~\cite{Kvasir_Seg}), which have 1000, 612, 196, and 1000 images, respectively. 
We followed~\cite{PraNet} to resize each image to $352\times 352$ and normalize the resized image by the statistics computed on ImageNet. 
The $DSC$, enhanced-alignment metric ($E^{max}_\phi$)~\cite{Metric_E_max}, and structural similarity metric ($S_\alpha$)~\cite{Metric_S_alpha} are utilized for evaluation in this task.

\subsection{Implementation Details}
For each task, we trained the source model on each single domain (source domain) and tested it on the left domains (target domains) to calculate the mean metrics to evaluate the methods on different scenarios.
In the source-training phase, we trained a ResUNet-34~\cite{ResNet} backbone, following~\cite{ProSFDA}, as the baseline for OD/OC segmentation task and trained the PraNet~\cite{PraNet} with a Res2Net-based~\cite{Res2Net} backbone for polyp segmentation task. Due to the complexity of PraNet's decoder, we only calculated the prompt loss $\mathcal{L}_p$ for each BN layer in the encoder instead of the whole network.
In the test-adaptation phase, We performed one-iteration adaptation for each batch of test data with a batch size of 1 on all experiments of our VPTTA and other competing methods~\cite{DLTTA}. 
As our VPTTA involves an iteration to update and an inference step, distinguishing it from other methods like~\cite{TENT} that do not require a dedicated inference step, we applied the same configuration as VPTTA to other competing methods to ensure consistency in the experimental setup.
To deploy our VPTTA, we utilized the Adam optimizer with a learning rate of 0.05 and 0.01 for the OD/OC segmentation task and polyp segmentation task, respectively. The hyperparameters $\alpha$ (size of prompt), $S$ (size of memory bank), $K$ (size of support set), and $\tau$ (temperature coefficient in warm-up) are set to 0.01, 40, 16, and 5 for both segmentation tasks.


\subsection{Experimental Results}
We compare our VPTTA with the `Source Only' baseline (training on the source domain and testing without adaptation), and six competing methods, including a pseudo-label-based method (CoTTA~\cite{CoTTA}), two entropy-based methods (TENT-continual~\cite{TENT} and SAR~\cite{SAR}), a method dynamically adjusting the learning rate (DLTTA~\cite{DLTTA}), a method combining entropy-loss and BN statistics fusion (DomainAdaptor~\cite{DomainAdaptor}), and a BN statistics-based method (DUA~\cite{DUA}) on two segmentation tasks. Specifically, we re-implemented all the competing methods using the same baseline as VPTTA and combined DLTTA with the entropy loss proposed in TENT~\cite{DLTTA}. 
The result of each domain is calculated by using the current domain as the source domain and applying the pre-trained model on the left domains.

\noindent \textbf{Comparison on the OD/OC segmentation task.}
The results of the proposed VPTTA, the “Source Only” baseline, and six competing methods on the OD/OC segmentation task are detailed in Table~\ref{tab:Comparison_OD/OC}. 
Obviously, all competing methods outperform the “Source Only” baseline, indicating the effectiveness of adaptation towards the distribution shifts.
It can be found that our VPTTA achieves the best overall performance across all domains, underscoring its superior applicability and robustness. Meanwhile, our VPTTA also shows competitive performance in domains where most methods exhibit satisfactory performance (such as D) and also performs well in domains where other methods exhibit suboptimal results (such as C).

\begin{figure*}[!htb]
   \centering
   \includegraphics[width=0.95\textwidth]{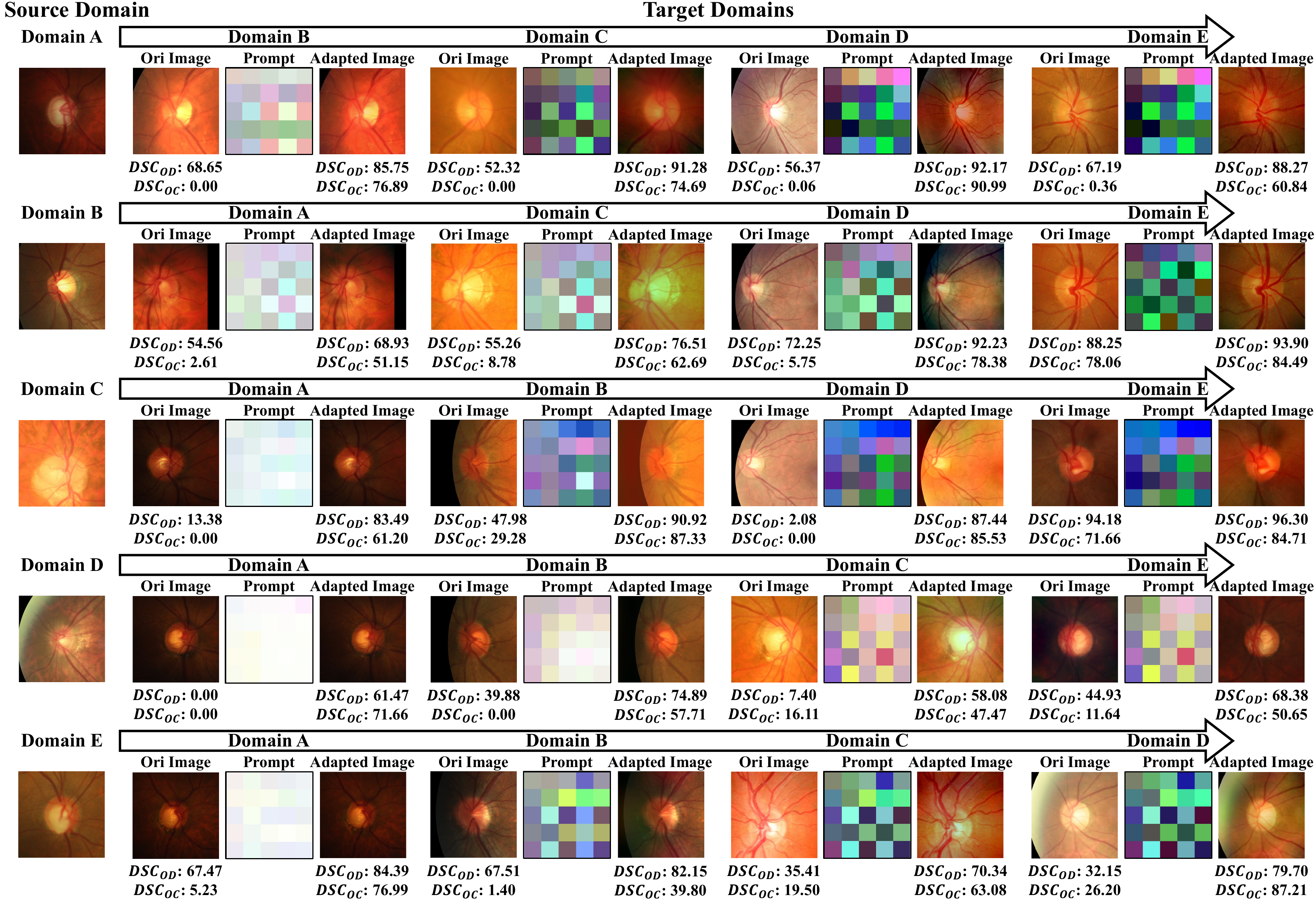}
   \caption{Visualization of the original images, estimated prompts, and adapted images on the OD/OC segmentation task. We normalize the prompts to [0, 1] for better visualization. The DSC of applying the frozen source model on the original and adapted images is displayed below each image. We also show an example of each source domain on the left side of this diagram. `Ori': Abbreviation of `Original'.}   
   \label{fig:Visualization}
\end{figure*}

\noindent \textbf{Comparison on the polyp segmentation task.}
We conducted the comparison experiment on the polyp segmentation task under the same experimental setup, as shown in Table~\ref{tab:Polyp}.
In contrast to the results of the OD/OC segmentation task, the methods based on model updating (\emph{i.e.}, TENT-continual, CoTTA, DLTTA, SAR, and DomainAdaptor) exhibit significantly worse performance in this task and are even inferior to the `Source Only' baseline.
It can be attributed to the fact that the polyps are relatively hidden, and the models tend to segment almost nothing rather than poorly segment under the impact of distribution shifts, resulting in confident and low entropy predictions but wrong gradients for these methods.
Since DUA is backward-free, it still performs well in this task.
However, DUA also fails when using domain A as the source domain and has limited performance due to the absence of training for the current test image.
Our VPTTA simultaneously benefits from training on the current test image and refraining from updating the model, thus surpassing all competing methods and demonstrating advanced performance.

\noindent \textbf{Visualization of prompts and adapted images.}
We visualized the estimated prompts and adapted images on the OD/OC segmentation task in Figure~\ref{fig:Visualization} and displayed the DSC values predicted on the original and adapted images below the corresponding images.
It can be observed that: (1) Applying prompts to the original test images can achieve significant performance gain; (2) Images from different target domains have similar prompts, proving that involving the memory bank to initialize the current prompt using previous ones is reasonable; (3) The appearances of adapted images are close to the source domain, explicitly demonstrating the reduction of distribution shifts via our VPTTA.

\begin{table*}[!ht]
    \caption{Performance of our VPTTA and four competing methods on the OD/OC segmentation task in a long-term continual test-time adaptation. 
    The \textcolor[rgb]{1,0,0}{red} numbers indicate the performance gain relative to the `Source Only' baseline. The performance degradation is calculated between the overall average DSC and the average DSC of round 1. `Ave': Abbreviation of `Average'.
    }
    \centering
    \resizebox{1\textwidth}{!}{
    \begin{tabular}{c|cccccc|cccccc|cccccc|cc}
        \Xhline{1pt}
        Time &
        \multicolumn{20}{c}{$\xrightarrow{\hspace{26cm}}$} \\
        \hline
        Round &
        \multicolumn{6}{c|}{1} &
        \multicolumn{6}{c|}{2} &
        \multicolumn{6}{c|}{3} &
        Average & Perform. \\
        \Xcline{1-19}{0.4pt}
        Methods & 
        A & B & C & D & E & Ave &
        A & B & C & D & E & Ave &
        A & B & C & D & E & Ave &
        DSC $\uparrow$ & Degra. $\downarrow$\\ 
        \hline

        Source Only &
        $64.53$ & $76.06$ & $71.18$ & $52.67$ & $64.87$ & $65.86$ &  
        $64.53$ & $76.06$ & $71.18$ & $52.67$ & $64.87$ & $65.86$ &  
        $64.53$ & $76.06$ & $71.18$ & $52.67$ & $64.87$ & $65.86$ &  
        $65.86$ & - \\
        \hline

        TENT &
        $73.07$ & $78.66$ & $71.94$ & $46.81$ & $70.20$ & $68.13$ & 
        $62.09$ & $69.32$ & $70.67$ & $39.02$ & $68.22$ & $61.86$ & 	
        $57.05$ & $62.47$ & $70.20$ & $39.02$ & $66.37$ & $59.02$ & 
        $63.01$ (\textcolor[rgb]{1,0,0}{$-2.85$}) & $5.12$\\ 

        CoTTA &
        $75.39$ & $75.98$ & $69.14$ & $53.99$ & $70.40$ & $68.98$ & 	
        $74.31$ & $75.00$ & $67.99$ & $51.04$ & $68.28$ & $67.32$ & 	
        $73.22$ & $74.33$ & $66.72$ & $50.23$ & $67.08$ & $66.32$ & 	
        $67.54$ (\textcolor[rgb]{1,0,0}{$+1.68$}) & $1.44$\\ 

        DLTTA &
        $75.11$ & $78.85$ & $73.89$ & $51.64$ & $69.71$ & $69.84$ & 	
        $74.14$ & $79.65$ & $74.25$ & $45.05$ & $69.04$ & $68.43$ & 
        $72.28$ & $78.93$ & $72.87$ & $42.37$ & $69.26$ & $67.14$ & 	
        $68.47$ (\textcolor[rgb]{1,0,0}{$+2.61$}) & $1.37$\\

        SAR &
        $74.55$ & $77.71$ & $70.78$ & $55.40$ & $71.72$ & $70.03$ & 	
        $74.74$ & $78.09$ & $71.00$ & $52.13$ & $69.02$ & $69.00$ & 	
        $74.90$ & $78.24$ & $71.18$ & $50.16$ & $68.44$ & $68.58$ & 	
        $69.20$ (\textcolor[rgb]{1,0,0}{$+3.34$}) & $0.83$\\ 
        \hline

        VPTTA (Ours) &
        $73.91$ & $79.36$ & $74.51$ & $56.51$ & $75.35$ & $71.93$ & 	
        $73.57$ & $78.84$ & $73.61$ & $56.91$ & $74.80$ & $71.55$ & 	
        $73.12$ & $78.45$ & $72.63$ & $57.11$ & $74.04$ & $71.07$ & 	
        $71.51$ (\textcolor[rgb]{1,0,0}{$+5.65$}) & $0.42$\\ 
        \hline
        \Xhline{1pt}
    \end{tabular}
    }
    \label{tab:continual}
\end{table*}

\noindent \textbf{Comparison in a long-term continual test-time adaptation.}
To evaluate our VPTTA in a long-term continual test-time adaptation~\cite{CoTTA}, we conducted the experiments for the OD/OC segmentation task multiple rounds and reported the results in Table~\ref{tab:continual}.  
Since DUA is a backward-free method and DomainAdaptor resets the pre-trained model before each adaptation, they are unsuitable to be evaluated in this experiment.
We compared our VPTTA with TENT, CoTTA, DLTTA, and SAR, all of which involve continuous training of the pre-trained model.  
The performance across different rounds with the same source domain reflects the extent of catastrophic forgetting in the model, and the average performance across multiple rounds highlights the error accumulation of the model within a series of distinct domains.
The results reveal that both optimizing loss functions (CoTTA, DLTTA, SAR) and resetting the model (CoTTA, SAR) are effective for mitigating error accumulation and catastrophic forgetting, while TENT with vanilla entropy-minimization loss suffers from severe performance degradation. 
Additionally, after multiple rounds of testing, our VPTTA maintains superior performance with minimal performance degradation, emphasizing the effectiveness of training prompts over updating the pre-trained model.

\noindent \textbf{Ablation study.}
To evaluate the contributions of the low-frequency prompt, memory-bank-based initialization, and warm-up mechanism, we conducted ablation studies on the OD/OC segmentation task, as shown in Table~\ref{tab:ablation}.
Note that Eq.~(\ref{eq3})/(\ref{eq5}) is used as the loss function when training the prompt without/with our warm-up mechanism.
The results show that (1) using the low-frequency prompt only has limited performance gain due to the poorly trained prompt; (2) both the memory bank and warm-up mechanism are beneficial to training the prompt; (3) the best performance is achieved when the memory bank and warm-up mechanism are jointly used, which means they are complementary.

\begin{figure}[!ht]
   \centering
   \includegraphics[width=0.98\columnwidth]{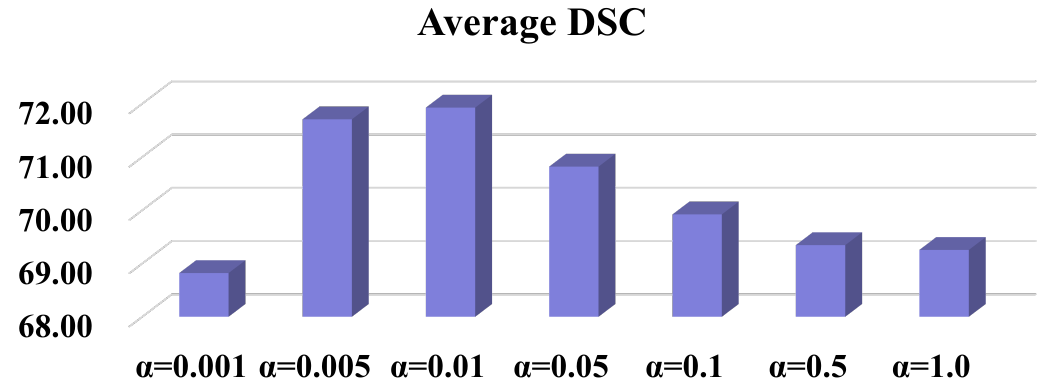}
   \caption{Performance of our VPTTA with various $\alpha$ on the OD/OC segmentation task.}
   \label{fig:alpha}
\end{figure}

\begin{figure}[!ht]
   \centering
   \includegraphics[width=0.98\columnwidth]{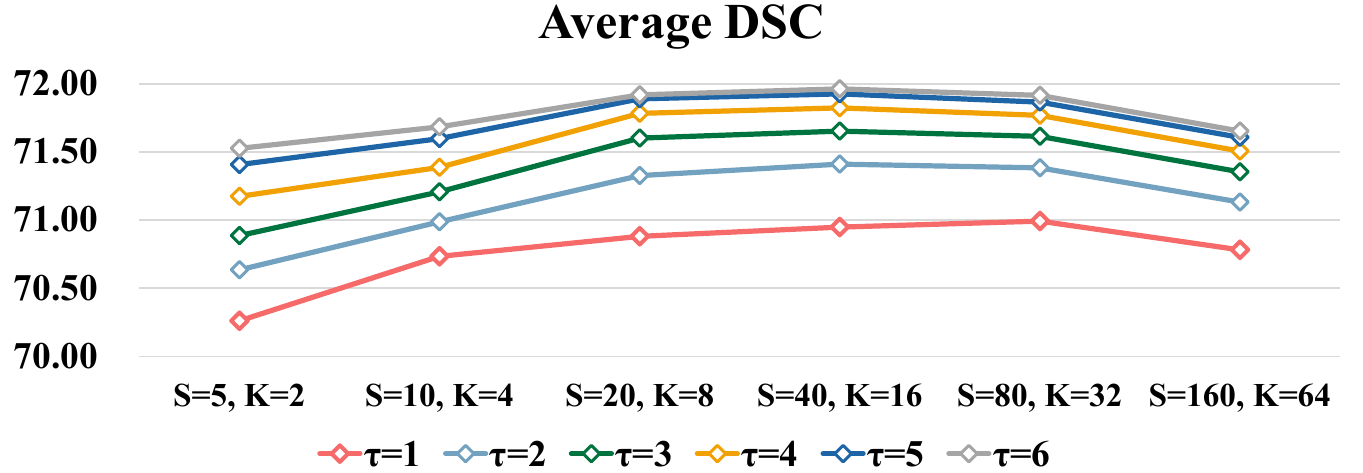}
   \caption{Performance of our VPTTA with various $S$, $K$, and $\tau$ on the OD/OC segmentation task.}
   \label{fig:grid}
\end{figure}



\noindent \textbf{Analysis of hyperparameter $\bm{\alpha}$, $\bm{S}$, $\bm{K}$, and $\bm{\tau}$.}
We conducted the experiments for the OD/OC segmentation task to discuss the hyper-parameters used in our VPTTA and displayed the results in Figure~\ref{fig:alpha} and Figure~\ref{fig:grid}.
It can be seen that $\alpha=0.01$ is optimal, as an excessively small prompt makes adaptation challenging, while an overly large prompt hinders effective training. 
Similar trends are also reflected in $S$ and $K$: as they increase, the DSC values improve, while the excessively large $S$ and $K$ may introduce unrelated samples, affecting the initialization. 
With the increase of $\tau$, performance gradually approaches the upper bound and tends to remain stable, suggesting that the warm-up has reached saturation.
Therefore, we finally chose $\alpha=0.01$, $S=40$, $K=16$, and $\tau=5$ as the best configuration.

\begin{table}[!tb]
    \caption{Performance of our VPTTA and VPTTA-LR on two segmentation tasks. 
    }
    \centering
    \resizebox{0.98\columnwidth}{!}{
    \begin{tabular}{c|cc|cccc}
        \Xhline{1pt}
        {\multirow{2}*{Methods}} & \multicolumn{2}{c|}{OD/OC} & \multicolumn{4}{c}{Polyp} \\
        \Xcline{2-7}{0.4pt}
         & $DSC\uparrow$ & Param $\downarrow$ & $DSC\uparrow$ & $E^{max}_\phi \uparrow$ & $S_\alpha \uparrow$ & Param $\downarrow$\\
        \hline
        VPTTA-LR (r=3) & $69.01$ & 9,216 & $78.25$ & $88.27$ & $84.20$ & 6,336\\
        VPTTA & $71.93$ & 75 & $80.46$ & $89.94$ & $86.84$ & 27\\

        \Xhline{1pt}
    \end{tabular}
    }
    \label{tab:vs}
\end{table}

\noindent \textbf{Low-frequency prompt vs. low-rank prompt.}
To compare the effectiveness of VPTTA and VPTTA-LR, we repeated the experiments on two segmentation tasks and showed results in Table~\ref{tab:vs}. It can be seen that VPTTA significantly outperforms VPTTA-LR, attributed to the fewer learnable parameters and superior representation of style texture in the frequency domain.
Benefiting from the low-frequency prompt, VPTTA can be effectively trained with over a hundred times fewer parameters than VPTTA-LR.


\section{Conclusion}
\label{sec:conclu}
In this paper, we proposed the VPTTA, a prompt-based method to alleviate the potential risks (\emph{i.e.}, error accumulation, and catastrophic forgetting) by freezing the pre-trained model. Specially, we presented the low-frequency prompt that trains a lightweight prompt with only a small number of parameters in a single iteration to alleviate the distribution shifts in BN layers, constructed a memory bank to better initialize each prompt by utilizing the prompts of previous and the most similar test images, and designed the statistics-fusion-based warm-up mechanism to simulate the warm-up statistics to overcome the training difficulty at the beginning of the inference phase.
The extensive experimental results on two medical image segmentation benchmark tasks with multiple domains verified the superiority of our VPTTA over other methods and the effectiveness of each special design.
In this regard, we hope that our efforts can offer a different paradigm for continual test-time adaptation.

\newpage

{
    \small
    \bibliographystyle{ieeenat_fullname}
    \bibliography{references}
}

\clearpage
\setcounter{page}{1}
\maketitlesupplementary
\appendix

\begin{figure*}[!htb]
   \centering
   \includegraphics[width=\textwidth]{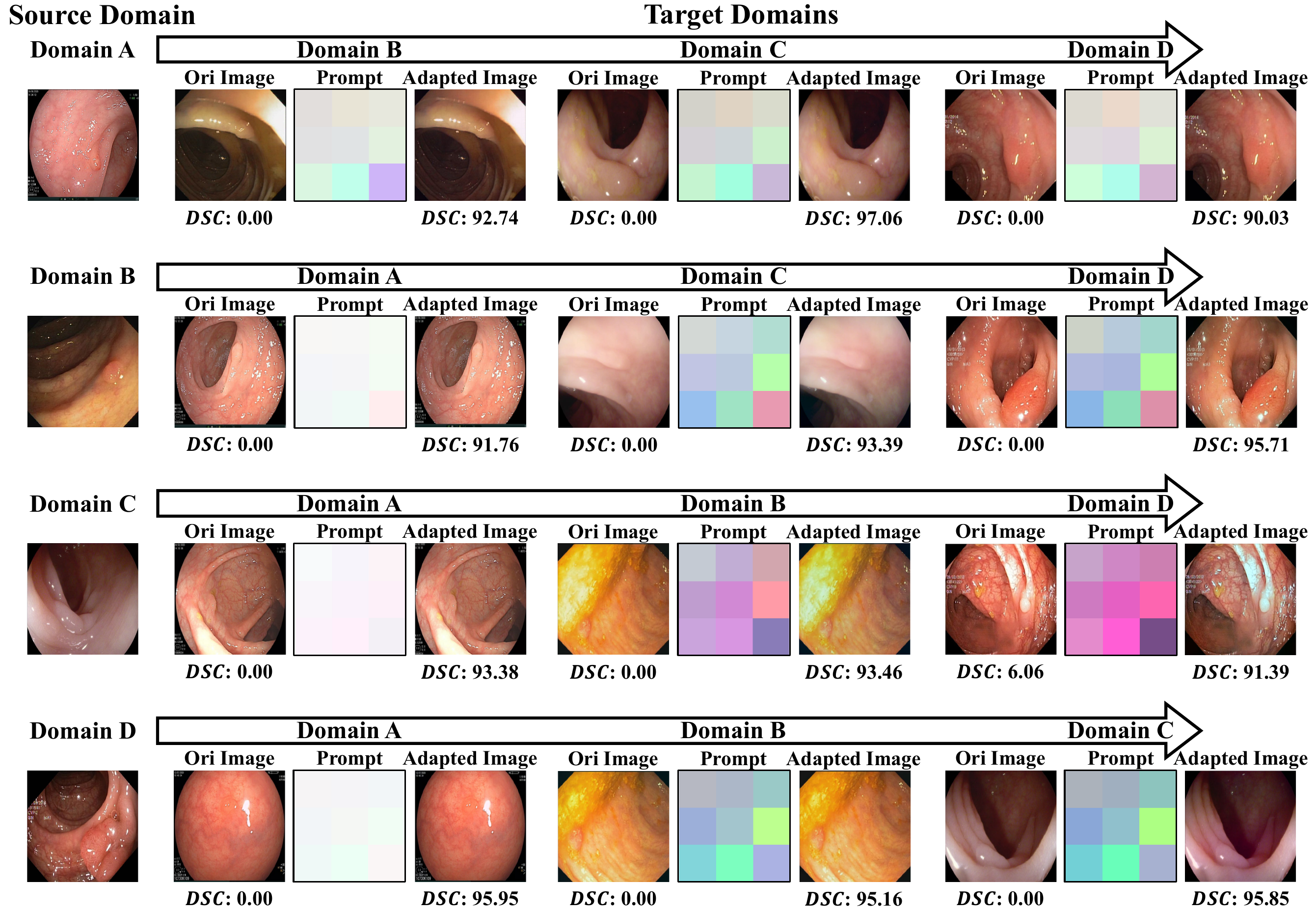}
   \caption{Visualization of the original images, estimated prompts, and adapted images on the polyp segmentation task. We normalize the prompts to [0, 1] for better visualization. The DSC of applying the frozen source model on the original and adapted images is displayed below each image. We also show an example of each source domain on the left side of this diagram. 'Ori': Abbreviation of 'Original'.}   
   \label{fig:Visualization_polyp}
\end{figure*}

\begin{table*}[!htb]
    \caption{Performance of our VPTTA, 'Source Only' baseline, and six competing methods on the OD/OC segmentation task. The best and second-best results in each column are highlighted in \textbf{bold} and \underline{underline}, respectively.
    }
    \centering
    \resizebox{0.9\textwidth}{!}{
    \begin{tabular}{c|ccccc|c}
        \Xhline{1pt}
        {\multirow{2}*{Methods}} & 
        \multicolumn{1}{c}{Domain A} & 
        \multicolumn{1}{c}{Domain B} & 
        \multicolumn{1}{c}{Domain C} & 
        \multicolumn{1}{c}{Domain D} & 
        \multicolumn{1}{c|}{Domain E} & 
        Average \\ 
        
        \Xcline{2-7}{0.4pt}
         &
        $DSC$ &
        $DSC$ &
        $DSC$ &
        $DSC$ &
        $DSC$ &
        $DSC\uparrow$ \\
        \hline

         Source Only (ResUNet-34) &
         $64.53$ & 
         $76.06$ & 
         $71.18$ & 
         $52.67$ &
         $64.87$ &
         $65.86$ \\
         \hline

         TENT-continual (ICLR 2021)~\cite{TENT} &
         $71.50$ & 
         $77.96$ & 
         $72.79$ & 
         $42.97$ &
         $69.56$ & 
         $66.96$ \\

         CoTTA (CVPR 2022)~\cite{CoTTA} &
         $73.71$ &
         $76.31$ &
         $72.43$ &	
         $53.04$ &
         $71.14$ &
         $69.33$ \\

         DLTTA (TMI 2022)~\cite{DLTTA} &
         $\textbf{74.90}$ & 	
         $\underline{78.73}$ & 	
         $\textbf{74.48}$ & 	
         $50.99$ & 	
         $69.25$ & 	
         $69.67$ \\ 

         DUA (CVPR 2022)~\cite{DUA} &
         $73.06$ & 
         $75.74$ & 
         $70.82$ & 
         $\underline{57.04}$ & 
         $70.31$ & 
         $69.39$ \\

         SAR (ICLR 2023)~\cite{SAR} & 
         $74.48$ &
         $77.49$ &
         $70.78$ & 		
         $\textbf{57.93}$ &
         $\underline{73.05}$ &
         $\underline{70.75}$ \\ 

         DomainAdaptor (CVPR 2023)~\cite{DomainAdaptor} &
         $\underline{74.50}$ &
         $76.39$ &
         $71.81$ &
         $56.78$ &
         $70.55$ & 
         $70.01$ \\
         \hline

         VPTTA (Ours) & 
         $74.24$ & 
         $\textbf{79.12}$ & 
         $\underline{74.05}$ & 
         $55.84$ & 
         $\textbf{76.47}$ &
         $\textbf{71.94}$ \\
        \Xhline{1pt}
    \end{tabular}
    }
    \label{tab:Mix_OD/OC}
\end{table*}

\begin{table*}[!htb]
    \caption{Performance of our VPTTA, 'Source Only' baseline, and six competing methods on the polyp segmentation task. The best and second-best results in each column are highlighted in \textbf{bold} and \underline{underline}, respectively.
    }
    \centering
    \resizebox{1\textwidth}{!}{
    \begin{tabular}{c|ccc ccc ccc ccc|ccc}
        \Xhline{1pt}
        {\multirow{2}*{Methods}} & 
        \multicolumn{3}{c}{Domain A} & 
        \multicolumn{3}{c}{Domain B} & 
        \multicolumn{3}{c}{Domain C} & 
        \multicolumn{3}{c|}{Domain D} & 
        \multicolumn{3}{c}{Average} \\ 
        \Xcline{2-16}{0.4pt}
         &
        $DSC$ & $E^{max}_\phi$ & $S_\alpha$ &
        $DSC$ & $E^{max}_\phi$ & $S_\alpha$ &
        $DSC$ & $E^{max}_\phi$ & $S_\alpha$ &
        $DSC$ & $E^{max}_\phi$ & $S_\alpha$ &
        $DSC \uparrow$ & $E^{max}_\phi \uparrow$ & $S_\alpha \uparrow$ \\
        \hline

         Source Only (PraNet) &
         $\underline{79.90}$ & $\underline{87.97}$ & $\underline{84.66}$ &
         $66.33$ & $78.51$ & $76.72$ & 	
         $73.89$ & $84.64$ & $81.28$ & 	
         $82.95$ & $90.84$ & $88.08$ & 	
         $75.77$ & $85.49$ & $82.69$\\
         \hline

         TENT-continual (ICLR 2021)~\cite{TENT} &
         $72.72$ & $82.99$ & $79.19$ & 	
         $69.41$ & $80.09$ & $79.10$ & 	
         $13.38$ & $36.09$ & $51.23$ & 	
         $73.70$ & $83.33$ & $82.72$ & 	
         $57.30$ & $70.62$ & $73.06$ \\

         CoTTA (CVPR 2022)~\cite{CoTTA} &
         $76.29$ & $85.31$ & $82.51$ & 	
         $66.58$ & $76.73$ & $79.11$ & 	
         $71.29$ & $83.50$ & $80.12$ & 	
         $70.62$ & $79.81$ & $82.56$ & 	
         $71.20$ & $81.34$ & $81.07$ \\ 

         DLTTA (TMI 2022)~\cite{DLTTA} &
         $75.52$ & $84.69$ & $81.88$ & 	
         $66.66$ & $77.21$ & $79.34$ & 	
         $63.75$ & $78.79$ & $75.55$ & 	
         $70.79$ & $81.14$ & $83.32$ & 	
         $69.18$ & $80.46$ & $80.02$ \\

         DUA (CVPR 2022)~\cite{DUA} &
         $78.79$ & $87.14$ & $83.93$ & 	
         $69.13$ & $80.62$ & $79.03$ & 	
         $\underline{74.66}$ & $\underline{84.96}$ & $\underline{82.07}$ & 	
         $\underline{86.63}$ & $\underline{93.62}$ & $\underline{90.06}$ & 	
         $\underline{77.30}$ & $\underline{86.58}$ & $\underline{83.77}$ \\

         SAR (ICLR 2023)~\cite{SAR} &
         $76.48$ & $85.89$ & $81.49$ & 	
         $66.45$ & $77.35$ & $78.05$ & 	
         $71.46$ & $83.23$ & $79.40$ & 	
         $70.41$ & $80.11$ & $81.07$ & 	
         $71.20$ & $81.65$ & $80.00$ \\

         DomainAdaptor (CVPR 2023)~\cite{DomainAdaptor} &
         $77.48$ & $86.31$ & $82.40$ & 	
         $\underline{70.82}$ & $\underline{81.76}$ & $\underline{80.88}$ & 	
         $71.96$ & $83.06$ & $79.97$ & 	
         $76.89$ & $85.89$ & $84.45$ & 	
         $74.29$ & $84.26$ & $81.93$ \\ 
         \hline

         VPTTA (Ours) & 
         $\textbf{80.65}$ & $\textbf{88.62}$ & $\textbf{84.78}$ & 
         $\textbf{76.94}$ & $\textbf{87.64}$ & $\textbf{84.10}$ & 
         $\textbf{76.48}$ & $\textbf{86.56}$ & $\textbf{83.04}$ & 
         $\textbf{86.37}$ & $\textbf{93.54}$ & $\textbf{89.87}$ & 
         $\textbf{80.11}$ & $\textbf{89.09}$ & $\textbf{85.45}$ \\
         \Xhline{1pt}
    \end{tabular}
    }
    \label{tab:Mix_Polyp}
\end{table*}

In the supplementary material, we provide more details about our implementation details of medical image segmentation tasks (Section~\ref{implement:med}) and natural classification image task (Section~\ref{implement:nat}), and additional experiments of medical image segmentation tasks (Section~\ref{experiments:med}) and natural image classification task (Section~\ref{experiments:nat}).

\section{Other Implementation Details of Medical Image Segmentation Tasks}
\label{implement:med}
\noindent \textbf{GPU device.} We conducted all experiments on a single RTX-2080Ti GPU for two medical image segmentation benchmark tasks. 

\noindent \textbf{Training the source model for the OD/OC segmentation task.} We utilized the SGD optimizer with a momentum of 0.99 and a weight decay of 0.0005, and the initial learning rate $lr_0$ was set to 0.001 and decayed according to $lr_t = lr_0 \times (1 - t/T)^{0.9}$, where $t$ is the current epoch and the maximum epoch $T$ is set to 200. The batch size was set to 8.
Empirically, we chose the model trained after the last epoch as the testing model.

\noindent \textbf{Training the source model for the polyp segmentation task.} We utilized the publicly released code of PraNet~\cite{PraNet}\footnote{\url{https://github.com/DengPingFan/PraNet}} to train the source model. 
The Adam algorithm with a learning rate of 0.0001 was adopted as the optimizer and the maximum epoch was set to 20. The batch size was set to 16.
Empirically, we chose the model trained after the last epoch as the testing model.

\noindent \textbf{Implementation details of different test-time adaptation methods.} Since these compared methods (\emph{i.e.}, TENT-continual~\cite{TENT}\footnote{\url{https://github.com/DequanWang/tent}}, CoTTA~\cite{CoTTA}\footnote{\url{https://github.com/qinenergy/cotta}}, DLTTA~\cite{DLTTA}\footnote{\url{https://github.com/med-air/DLTTA}}, DUA~\cite{DUA}\footnote{\url{https://github.com/jmiemirza/DUA}}, SAR~\cite{SAR}\footnote{\url{https://github.com/mr-eggplant/SAR}}, and DomainAdaptor~\cite{DomainAdaptor}\footnote{\url{https://github.com/koncle/DomainAdaptor}}) are all open-source, we adopted their source codes to conduct experiments. We set the batch size to 1 for all experiments following~\cite{DLTTA}.

\section{Other Implementation Details of Natural Image Classification Task}
\label{implement:nat}
\noindent \textbf{GPU device.} We conducted all experiments on a single RTX-3090 GPU.

\noindent \textbf{Datasets and Evaluation Metrics.} \textbf{PACS}~\cite{PACS} dataset is commonly used in domain generalization and test-time adaptation, which comprises 9,991 images and 7
classes that are collected from 4 distinct domains: art, cartoons, photos, and sketches. 

\noindent \textbf{Training the source model.} 
We trained the source model by empirical risk minimization (ERM)~\cite{ERM} algorithm with ResNet~\cite{ResNet} backbone using the SGD optimizer with a learning rate of 5e-5. The batch size was set to 32 and the number of training iteration was set to 5k. We resized all images to $224 \times 224$ and used data augmentation during training, including random cropping, random flipping, color jittering, and intensity changing.

\noindent \textbf{Implementation details of comparison experiments under the test-time adaptation setup.} 
We compared our VPTTA with four methods (\emph{i.e.}, BN~\cite{BN}, TENT~\cite{TENT}, SHOT-IM~\cite{SHOT-IM}, and TSD~\cite{TSD}) under the test-time adaptation setup following TSD (\emph{i.e.}, given D domains, training on D-1 domains and testing on the left one) and used the publicly released code of TSD\footnote{\url{https://github.com/SakurajimaMaiii/TSD}} to conduct experiments for all methods.
We resized all test images to $224 \times 224$ and no data augmentation was used.
For all experiments, we set the random seed to 0.
To deploy our VPTTA, we utilized the Adam optimizer with a learning rate of 0.01. The hyperparameters $\alpha$ (size of prompt), $S$ (size of memory bank), $K$ (size of support set), and $\tau$ (temperature coefficient in warm-up) are set to 0.02, 64, 4, and 1.
The code and weights of pre-trained source models are available at \href{https://github.com/Chen-Ziyang/VPTTA}{https://github.com/Chen-Ziyang/VPTTA}.

\section{More Experiments of Medical Image Segmentation Tasks}
\label{experiments:med}
\subsection{Visualization of prompts and adapted images on the polyp segmentation task}
We also visualized the prompts and adapted images on the polyp segmentation task, as shown in Figure~\ref{fig:Visualization_polyp}. 
We found that the prompts induce subtle alterations in the appearance of images, but still yield substantial performance gains, even on the hard samples which cannot be recognized by the model (\emph{i.e.}, $DSC=0.00$). 
Similar to the observation on the OD/OC segmentation task, the prompts of different target domains produced for the same source model exhibit high similarity.

\subsection{Comparison experiments under mixed distribution shifts}
Considering that test data may come arbitrarily in the complex real world, we conducted the experiments under the mixed distribution shifts, \emph{i.e.}, training the model on a single source domain and testing it on a mixed domain composed of the left target domains. We used 2024 as the random seed to shuffle the data of left target domains for all methods, and the batch size was set to 1. The results of two medical segmentation benchmark tasks are displayed in Table~\ref{tab:Mix_OD/OC} and Table~\ref{tab:Mix_Polyp}.
In Table~\ref{tab:Mix_Polyp}, we observed similar phenomena that only DUA and our VPTTA outperform the baseline, but other methods fail due to the wrong gradients produced by the confident but terrible predictions.
Meanwhile, the results in Table~\ref{tab:Mix_OD/OC} and Table~\ref{tab:Mix_Polyp} reveal that our VPTTA still achieves the best overall performance across all domains on both two tasks, underscoring its superior applicability and robustness. 

\begin{table}[!htb]
  \centering
  \caption{Performance of our VPTTA, ERM baseline, and four competing methods on the PACS dataset with ResNet-18 backbone. The best and
second-best results in each column are highlighted in bold and underline, respectively.}
  \resizebox{1\columnwidth}{!}{
    \begin{tabular}{c|c|ccccc}
    \Xhline{1pt}
    \multicolumn{2}{c|}{ResNet-18} & A     & C     & P     & S     & Average \\
    \hline
    \multicolumn{2}{c|}{ERM~\cite{ERM}} & 78.37  & 77.05  & 95.57  & 65.69  & 79.17  \\
    \hline
    \multirow{6}[0]{*}{BS=64} & BN (NeurIPS 2020)~\cite{BN} & 81.05  & 80.63  & 95.03  & 72.26  & 82.24  \\
          & Tent (ICLR 2021)~\cite{TENT} & 81.35  & 80.89  & 95.27  & 73.45  & 82.74  \\
          & SHOT-IM (ICML 2020)~\cite{SHOT-IM} & 82.71  & 76.75  & 94.97  & 67.32  & 80.44  \\
          & TSD (CVPR 2023)~\cite{TSD} & \textbf{87.89}  & \textbf{86.90}  & \textbf{96.53}  & \underline{71.77}  & \textbf{85.77}  \\
          & VPTTA (Ours) & \underline{81.15}  & \underline{80.67}  & \underline{96.23}  & \textbf{77.40}  & \underline{83.86}  \\
          \hline
    \multirow{6}[0]{*}{BS=1} & BN (NeurIPS 2020)~\cite{BN} & 13.09  & \underline{16.60}  & 11.32  & 5.17  & 11.54  \\
          & Tent (ICLR 2021)~\cite{TENT} & 10.06  & \underline{16.60}  & 11.32  & 4.07  & 10.51  \\
          & SHOT-IM (ICML 2020)~\cite{SHOT-IM} & \underline{13.77}  & \underline{16.60}  & 11.32  & 5.83  & 11.88  \\
          & TSD (CVPR 2023)~\cite{TSD} & 12.99  & 15.53  & \underline{11.80}  & \underline{16.75}  & \underline{14.27}  \\
          & VPTTA (Ours) & \textbf{79.35}  & \textbf{77.52} & \textbf{95.63}  & \textbf{69.56} & \textbf{80.51}  \\
    \Xhline{1pt}
    \end{tabular}%
}
  \label{tab:PACS_res18}%
\end{table}%

\begin{table}[!htb]
  \centering
  \caption{Performance of our VPTTA, ERM baseline, and four competing methods on the PACS dataset with ResNet-50 backbone. The best and
second-best results in each column are highlighted in bold and underline, respectively.}
  \resizebox{1\columnwidth}{!}{
    \begin{tabular}{c|c|ccccc}
    \Xhline{1pt}
    \multicolumn{2}{c|}{ResNet-50} & A     & C     & P     & S     & Average \\
    \Xhline{1pt}
          \multicolumn{2}{c|}{ERM~\cite{ERM}}   & 83.35  & 79.82  & 96.77  & 81.85  & 85.45  \\
    \hline
    \multirow{6}[0]{*}{BS=64} & BN (NeurIPS 2020)~\cite{BN} & 84.96  & 82.81  & 96.83  & 75.46  & 85.01  \\
          & Tent (ICLR 2021)~\cite{TENT} & 85.45  & 83.36  & 96.77  & 77.30  & 85.72  \\
          & SHOT-IM (ICML 2020)~\cite{SHOT-IM} & 83.30  & 82.00  & 93.35  & 63.99  & 80.66  \\
          & TSD (CVPR 2023)~\cite{TSD} & \textbf{88.87}  & \textbf{89.12}  & \textbf{97.43}  & \underline{82.13}  & \textbf{89.39}  \\
          & VPTTA (Ours) & \underline{86.47}  & \underline{83.53}  & \underline{97.37}  & \textbf{84.78}  & \underline{88.04}  \\
          \hline

    \multirow{6}[0]{*}{BS=1} & BN (NeurIPS 2020)~\cite{BN} & {29.30}  & \underline{16.60}  & 11.32  & 4.10  & 15.33  \\
          & Tent (ICLR 2021)~\cite{TENT} & 19.19  & \underline{16.60}  & 11.32  & 4.07  & 12.79  \\
          & SHOT-IM (ICML 2020)~\cite{SHOT-IM} & \underline{31.74}  & \underline{16.60}  & 11.32  & 4.20  & \underline{15.96}  \\
          & TSD (CVPR 2023)~\cite{TSD} & 12.21  & 14.16  & \underline{12.57}  & \underline{17.03}  & 13.99  \\
          & VPTTA (Ours) & \textbf{84.57}  & \textbf{81.66}  & \textbf{97.25}  & \textbf{82.62}  & \textbf{86.52}  \\
          \Xhline{1pt}

        \end{tabular}%
    }
  \label{tab:PACS_res50}%
\end{table}%



\section{More Experiments of Natural Image Classification Task}
\label{experiments:nat}
We evaluated our VPTTA and four methods with different batch sizes (BS) on the PACS dataset. The results are shown in Table~\ref{tab:PACS_res18} and Table~\ref{tab:PACS_res50}.
We found that other compared methods, such as TSD, perform well with a large test batch size (BS=64) but fail with a small test batch size (BS=1), which means they heavily depend on batch size.
The results show that our VPTTA surpasses other compared methods with a small test batch size and achieves competitive performance with a large test batch size.
It demonstrates that our VPTTA is suitable to be deployed in scenarios with small batch sizes.


\end{document}

%% file: preamble.tex
\usepackage[dvipsnames]{xcolor}


%% file: VPTTA_Arxiv.bbl
\begin{thebibliography}{57}
\providecommand{\natexlab}[1]{#1}
\providecommand{\url}[1]{\texttt{#1}}
\expandafter\ifx\csname urlstyle\endcsname\relax
  \providecommand{\doi}[1]{doi: #1}\else
  \providecommand{\doi}{doi: \begingroup \urlstyle{rm}\Url}\fi

\bibitem[Bernal et~al.(2015)Bernal, S{\'a}nchez, Fern{\'a}ndez-Esparrach, Gil, Rodr{\'\i}guez, and Vilari{\~n}o]{CVC_ClinicDB}
Jorge Bernal, F~Javier S{\'a}nchez, Gloria Fern{\'a}ndez-Esparrach, Debora Gil, Cristina Rodr{\'\i}guez, and Fernando Vilari{\~n}o.
\newblock Wm-dova maps for accurate polyp highlighting in colonoscopy: Validation vs. saliency maps from physicians.
\newblock \emph{Comput. Med. Imaging and Graph.}, 43:\penalty0 99--111, 2015.

\bibitem[Cascante-Bonilla et~al.(2021)Cascante-Bonilla, Tan, Qi, and Ordonez]{Pseudo2}
Paola Cascante-Bonilla, Fuwen Tan, Yanjun Qi, and Vicente Ordonez.
\newblock Curriculum labeling: Revisiting pseudo-labeling for semi-supervised learning.
\newblock In \emph{AAAI}, pages 6912--6920, 2021.

\bibitem[Chang et~al.(2019)Chang, Flokas, and Lipson]{Initialize2}
Oscar Chang, Lampros Flokas, and Hod Lipson.
\newblock Principled weight initialization for hypernetworks.
\newblock In \emph{Int. Conf. Learn. Represent.}, 2019.

\bibitem[Chen et~al.(2022{\natexlab{a}})Chen, Wang, Darrell, and Ebrahimi]{Contrastive}
Dian Chen, Dequan Wang, Trevor Darrell, and Sayna Ebrahimi.
\newblock Contrastive test-time adaptation.
\newblock In \emph{IEEE Conf. Comput. Vis. Pattern Recog.}, pages 295--305, 2022{\natexlab{a}}.

\bibitem[Chen et~al.(2022{\natexlab{b}})Chen, Chen, Wei, Jin, Tan, Jin, and Chen]{DALN}
Lin Chen, Huaian Chen, Zhixiang Wei, Xin Jin, Xiao Tan, Yi Jin, and Enhong Chen.
\newblock Reusing the task-specific classifier as a discriminator: Discriminator-free adversarial domain adaptation.
\newblock In \emph{IEEE Conf. Comput. Vis. Pattern Recog.}, pages 7181--7190, 2022{\natexlab{b}}.

\bibitem[D{\"o}bler et~al.(2023)D{\"o}bler, Marsden, and Yang]{RMT}
Mario D{\"o}bler, Robert~A Marsden, and Bin Yang.
\newblock Robust mean teacher for continual and gradual test-time adaptation.
\newblock In \emph{IEEE Conf. Comput. Vis. Pattern Recog.}, pages 7704--7714, 2023.

\bibitem[Fan et~al.(2017)Fan, Cheng, Liu, Li, and Borji]{Metric_S_alpha}
Deng-Ping Fan, Ming-Ming Cheng, Yun Liu, Tao Li, and Ali Borji.
\newblock Structure-measure: A new way to evaluate foreground maps.
\newblock In \emph{Int. Conf. Comput. Vis.}, pages 4548--4557, 2017.

\bibitem[Fan et~al.(2018)Fan, Gong, Cao, Ren, Cheng, and Borji]{Metric_E_max}
Deng-Ping Fan, Cheng Gong, Yang Cao, Bo Ren, Ming-Ming Cheng, and Ali Borji.
\newblock Enhanced-alignment measure for binary foreground map evaluation.
\newblock In \emph{IJCAI}, pages 698--704, 2018.

\bibitem[Fan et~al.(2020)Fan, Ji, Zhou, Chen, Fu, Shen, and Shao]{PraNet}
Deng-Ping Fan, Ge-Peng Ji, Tao Zhou, Geng Chen, Huazhu Fu, Jianbing Shen, and Ling Shao.
\newblock Pranet: Parallel reverse attention network for polyp segmentation.
\newblock In \emph{Int. Conf. Med. Image Comput. Comput.-Assist. Intervent.}, pages 263--273. Springer, 2020.

\bibitem[Frigo and Johnson(1998)]{FFTW}
Matteo Frigo and Steven~G Johnson.
\newblock {FFTW}: An adaptive software architecture for the fft.
\newblock In \emph{ICASSP}, pages 1381--1384. IEEE, 1998.

\bibitem[Fumero et~al.(2011)Fumero, Alay{\'o}n, Sanchez, Sigut, and Gonzalez-Hernandez]{RIM_ONE_r3}
Francisco Fumero, Silvia Alay{\'o}n, Jos{\'e}~L Sanchez, Jose Sigut, and M Gonzalez-Hernandez.
\newblock {RIM-ONE}: An open retinal image database for optic nerve evaluation.
\newblock In \emph{Int. Symp. Comput.-based Med. Syst.}, pages 1--6. IEEE, 2011.

\bibitem[Gan et~al.(2023)Gan, Bai, Lou, Ma, Zhang, Shi, and Luo]{DPT}
Yulu Gan, Yan Bai, Yihang Lou, Xianzheng Ma, Renrui Zhang, Nian Shi, and Lin Luo.
\newblock Decorate the newcomers: Visual domain prompt for continual test time adaptation.
\newblock In \emph{AAAI}, pages 7595--7603, 2023.

\bibitem[Gandelsman et~al.(2022)Gandelsman, Sun, Chen, and Efros]{TTAMAE}
Yossi Gandelsman, Yu Sun, Xinlei Chen, and Alexei Efros.
\newblock Test-time training with masked autoencoders.
\newblock \emph{Adv. Neural Inform. Process. Syst.}, 35:\penalty0 29374--29385, 2022.

\bibitem[Gao et~al.(2019)Gao, Cheng, Zhao, Zhang, Yang, and Torr]{Res2Net}
Shang-Hua Gao, Ming-Ming Cheng, Kai Zhao, Xin-Yu Zhang, Ming-Hsuan Yang, and Philip Torr.
\newblock Res2net: A new multi-scale backbone architecture.
\newblock \emph{IEEE Trans. Pattern Anal. Mach. Intell.}, 43\penalty0 (2):\penalty0 652--662, 2019.

\bibitem[Gao et~al.(2022)Gao, Shi, Zhu, Wang, Tang, Zhou, Li, and Metaxas]{Prompt}
Yunhe Gao, Xingjian Shi, Yi Zhu, Hao Wang, Zhiqiang Tang, Xiong Zhou, Mu Li, and Dimitris~N Metaxas.
\newblock Visual prompt tuning for test-time domain adaptation.
\newblock \emph{arXiv preprint arXiv:2210.04831}, 2022.

\bibitem[Ghafoorian et~al.(2017)Ghafoorian, Mehrtash, Kapur, Karssemeijer, Marchiori, Pesteie, Guttmann, de~Leeuw, Tempany, Van~Ginneken, et~al.]{domain_shift_2}
Mohsen Ghafoorian, Alireza Mehrtash, Tina Kapur, Nico Karssemeijer, Elena Marchiori, Mehran Pesteie, Charles~RG Guttmann, Frank-Erik de Leeuw, Clare~M Tempany, Bram Van~Ginneken, et~al.
\newblock Transfer learning for domain adaptation in mri: Application in brain lesion segmentation.
\newblock In \emph{Int. Conf. Med. Image Comput. Comput.-Assist. Intervent.}, pages 516--524. Springer, 2017.

\bibitem[He et~al.(2016)He, Zhang, Ren, and Sun]{ResNet}
Kaiming He, Xiangyu Zhang, Shaoqing Ren, and Jian Sun.
\newblock Deep residual learning for image recognition.
\newblock In \emph{IEEE Conf. Comput. Vis. Pattern Recog.}, pages 770--778, 2016.

\bibitem[Hu et~al.(2022{\natexlab{a}})Hu, Wallis, Allen-Zhu, Li, Wang, Wang, Chen, et~al.]{LoRa}
Edward~J Hu, Phillip Wallis, Zeyuan Allen-Zhu, Yuanzhi Li, Shean Wang, Lu Wang, Weizhu Chen, et~al.
\newblock Lora: Low-rank adaptation of large language models.
\newblock In \emph{Int. Conf. Learn. Represent.}, 2022{\natexlab{a}}.

\bibitem[Hu et~al.(2022{\natexlab{b}})Hu, Liao, and Xia]{ProSFDA}
Shishuai Hu, Zehui Liao, and Yong Xia.
\newblock Prosfda: Prompt learning based source-free domain adaptation for medical image segmentation.
\newblock \emph{arXiv preprint arXiv:2211.11514}, 2022{\natexlab{b}}.

\bibitem[Jha et~al.(2020)Jha, Smedsrud, Riegler, Halvorsen, Lange, Johansen, and Johansen]{Kvasir_Seg}
Debesh Jha, Pia~H Smedsrud, Michael~A Riegler, P{\aa}l Halvorsen, Thomas~de Lange, Dag Johansen, and H{\aa}vard~D Johansen.
\newblock Kvasir-seg: A segmented polyp dataset.
\newblock In \emph{Int. Conf. Multimedia Modeling}, pages 451--462. Springer, 2020.

\bibitem[Lamers et~al.(2023)Lamers, Vidal, Belbachir, van Stein, B{\"a}eck, and Giampouras]{Forgetting}
Christiaan Lamers, Ren{\'e} Vidal, Nabil Belbachir, Niki van Stein, Thomas B{\"a}eck, and Paris Giampouras.
\newblock Clustering-based domain-incremental learning.
\newblock In \emph{Int. Conf. Comput. Vis.}, pages 3384--3392, 2023.

\bibitem[Lee et~al.(2013)Lee, Kim, Lebanon, and Singer]{Compression}
Joonseok Lee, Seungyeon Kim, Guy Lebanon, and Yoram Singer.
\newblock Local low-rank matrix approximation.
\newblock In \emph{Int. Conf. Mach. Learn.}, pages 82--90. PMLR, 2013.

\bibitem[Li et~al.(2017)Li, Yang, Song, and Hospedales]{PACS}
Da Li, Yongxin Yang, Yi-Zhe Song, and Timothy~M Hospedales.
\newblock Deeper, broader and artier domain generalization.
\newblock In \emph{Int. Conf. Comput. Vis.}, pages 5542--5550, 2017.

\bibitem[Liang et~al.(2019)Liang, He, Sun, and Tan]{Pseudo1}
Jian Liang, Ran He, Zhenan Sun, and Tieniu Tan.
\newblock Exploring uncertainty in pseudo-label guided unsupervised domain adaptation.
\newblock \emph{Pattern Recognition}, 96:\penalty0 106996, 2019.

\bibitem[Liang et~al.(2020)Liang, Hu, and Feng]{SHOT-IM}
Jian Liang, Dapeng Hu, and Jiashi Feng.
\newblock Do we really need to access the source data? source hypothesis transfer for unsupervised domain adaptation.
\newblock In \emph{Int. Conf. Mach. Learn.}, pages 6028--6039. PMLR, 2020.

\bibitem[Liu et~al.(2023)Liu, Yuan, Fu, Jiang, Hayashi, and Neubig]{NLP}
Pengfei Liu, Weizhe Yuan, Jinlan Fu, Zhengbao Jiang, Hiroaki Hayashi, and Graham Neubig.
\newblock Pre-train, prompt, and predict: A systematic survey of prompting methods in natural language processing.
\newblock \emph{ACM Comput. Surveys}, 55\penalty0 (9):\penalty0 1--35, 2023.

\bibitem[Mirza et~al.(2022)Mirza, Micorek, Possegger, and Bischof]{DUA}
M~Jehanzeb Mirza, Jakub Micorek, Horst Possegger, and Horst Bischof.
\newblock The norm must go on: Dynamic unsupervised domain adaptation by normalization.
\newblock In \emph{IEEE Conf. Comput. Vis. Pattern Recog.}, pages 14765--14775, 2022.

\bibitem[Mummadi et~al.(2021)Mummadi, Hutmacher, Rambach, Levinkov, Brox, and Metzen]{Transformation}
Chaithanya~Kumar Mummadi, Robin Hutmacher, Kilian Rambach, Evgeny Levinkov, Thomas Brox, and Jan~Hendrik Metzen.
\newblock Test-time adaptation to distribution shift by confidence maximization and input transformation.
\newblock \emph{arXiv preprint arXiv:2106.14999}, 2021.

\bibitem[Nado et~al.(2020)Nado, Padhy, Sculley, D'Amour, Lakshminarayanan, and Snoek]{BN}
Zachary Nado, Shreyas Padhy, D Sculley, Alexander D'Amour, Balaji Lakshminarayanan, and Jasper Snoek.
\newblock Evaluating prediction-time batch normalization for robustness under covariate shift.
\newblock \emph{arXiv preprint arXiv:2006.10963}, 2020.

\bibitem[Ngoc~Lan et~al.(2021)Ngoc~Lan, An, Hang, Long, Trung, Thuy, and Sang]{BKAI_IGH_NeoPolyp}
Phan Ngoc~Lan, Nguyen~Sy An, Dao~Viet Hang, Dao~Van Long, Tran~Quang Trung, Nguyen~Thi Thuy, and Dinh~Viet Sang.
\newblock Neounet: Towards accurate colon polyp segmentation and neoplasm detection.
\newblock In \emph{Adv. Vis. Comput. Int. Symp.}, pages 15--28. Springer, 2021.

\bibitem[Nguyen et~al.(2023)Nguyen, Nguyen-Tang, Lim, and Torr]{TIPI}
A~Tuan Nguyen, Thanh Nguyen-Tang, Ser-Nam Lim, and Philip~HS Torr.
\newblock Tipi: Test time adaptation with transformation invariance.
\newblock In \emph{IEEE Conf. Comput. Vis. Pattern Recog.}, pages 24162--24171, 2023.

\bibitem[Niu et~al.(2022)Niu, Wu, Zhang, Chen, Zheng, Zhao, and Tan]{EATA}
Shuaicheng Niu, Jiaxiang Wu, Yifan Zhang, Yaofo Chen, Shijian Zheng, Peilin Zhao, and Mingkui Tan.
\newblock Efficient test-time model adaptation without forgetting.
\newblock In \emph{Int. Conf. Mach. Learn.}, pages 16888--16905. PMLR, 2022.

\bibitem[Niu et~al.(2023)Niu, Wu, Zhang, Wen, Chen, Zhao, and Tan]{SAR}
Shuaicheng Niu, Jiaxiang Wu, Yifan Zhang, Zhiquan Wen, Yaofo Chen, Peilin Zhao, and Mingkui Tan.
\newblock Towards stable test-time adaptation in dynamic wild world.
\newblock In \emph{Int. Conf. Learn. Represent.}, 2023.

\bibitem[Orlando et~al.(2020)Orlando, Fu, Breda, van Keer, Bathula, Diaz-Pinto, Fang, Heng, Kim, Lee, et~al.]{REFUGE}
Jos{\'e}~Ignacio Orlando, Huazhu Fu, Jo{\~a}o~Barbosa Breda, Karel van Keer, Deepti~R Bathula, Andr{\'e}s Diaz-Pinto, Ruogu Fang, Pheng-Ann Heng, Jeyoung Kim, JoonHo Lee, et~al.
\newblock {REFUGE Challenge}: A unified framework for evaluating automated methods for glaucoma assessment from fundus photographs.
\newblock \emph{Med. Image Anal.}, 59:\penalty0 101570, 2020.

\bibitem[Pan et~al.(2018)Pan, Luo, Shi, and Tang]{IBN}
Xingang Pan, Ping Luo, Jianping Shi, and Xiaoou Tang.
\newblock Two at once: Enhancing learning and generalization capacities via ibn-net.
\newblock In \emph{Eur. Conf. Comput. Vis.}, pages 464--479, 2018.

\bibitem[Sankaranarayanan et~al.(2018)Sankaranarayanan, Balaji, Jain, Lim, and Chellappa]{domain_shift_cvpr}
Swami Sankaranarayanan, Yogesh Balaji, Arpit Jain, Ser~Nam Lim, and Rama Chellappa.
\newblock Learning from synthetic data: Addressing domain shift for semantic segmentation.
\newblock In \emph{IEEE Conf. Comput. Vis. Pattern Recog.}, pages 3752--3761, 2018.

\bibitem[Schneider et~al.(2020)Schneider, Rusak, Eck, Bringmann, Brendel, and Bethge]{BNMismatch}
Steffen Schneider, Evgenia Rusak, Luisa Eck, Oliver Bringmann, Wieland Brendel, and Matthias Bethge.
\newblock Improving robustness against common corruptions by covariate shift adaptation.
\newblock \emph{Adv. Neural Inform. Process. Syst.}, 33:\penalty0 11539--11551, 2020.

\bibitem[Silva et~al.(2014)Silva, Histace, Romain, Dray, and Granado]{ETIS_LaribPolypDB}
Juan Silva, Aymeric Histace, Olivier Romain, Xavier Dray, and Bertrand Granado.
\newblock Toward embedded detection of polyps in wce images for early diagnosis of colorectal cancer.
\newblock \emph{Int. J. Comput.-Assist. Rad. Sur.}, 9\penalty0 (2):\penalty0 283--293, 2014.

\bibitem[Sinha et~al.(2023)Sinha, Gehler, Locatello, and Schiele]{TeST}
Samarth Sinha, Peter Gehler, Francesco Locatello, and Bernt Schiele.
\newblock Test: Test-time self-training under distribution shift.
\newblock In \emph{IEEE Wint. Conf. Appli. Comput. Vis.}, pages 2759--2769, 2023.

\bibitem[Sivaswamy et~al.(2014)Sivaswamy, Krishnadas, Joshi, Jain, and Tabish]{Drishti-GS}
Jayanthi Sivaswamy, SR Krishnadas, Gopal~Datt Joshi, Madhulika Jain, and A~Ujjwaft~Syed Tabish.
\newblock {Drishti-GS}: Retinal image dataset for optic nerve head (onh) segmentation.
\newblock In \emph{IEEE Int. Symp. on Bio. Imaging}, pages 53--56. IEEE, 2014.

\bibitem[Song et~al.(2023)Song, Lee, Kweon, and Choi]{EcoTTA}
Junha Song, Jungsoo Lee, In~So Kweon, and Sungha Choi.
\newblock Ecotta: Memory-efficient continual test-time adaptation via self-distilled regularization.
\newblock In \emph{IEEE Conf. Comput. Vis. Pattern Recog.}, pages 11920--11929, 2023.

\bibitem[Sun et~al.(2020)Sun, Wang, Liu, Miller, Efros, and Hardt]{TTT}
Yu Sun, Xiaolong Wang, Zhuang Liu, John Miller, Alexei Efros, and Moritz Hardt.
\newblock Test-time training with self-supervision for generalization under distribution shifts.
\newblock In \emph{Int. Conf. Mach. Learn.}, pages 9229--9248. PMLR, 2020.

\bibitem[Vapnik(1998)]{ERM}
Vladimir Vapnik.
\newblock Statistical learning theory.
\newblock \emph{Wiley}, 1998.

\bibitem[Wang et~al.(2021)Wang, Shelhamer, Liu, Olshausen, and Darrell]{TENT}
Dequan Wang, Evan Shelhamer, Shaoteng Liu, Bruno Olshausen, and Trevor Darrell.
\newblock Tent: Fully test-time adaptation by entropy minimization.
\newblock In \emph{Int. Conf. Learn. Represent.}, 2021.

\bibitem[Wang et~al.(2022)Wang, Fink, Van~Gool, and Dai]{CoTTA}
Qin Wang, Olga Fink, Luc Van~Gool, and Dengxin Dai.
\newblock Continual test-time domain adaptation.
\newblock In \emph{IEEE Conf. Comput. Vis. Pattern Recog.}, pages 7201--7211, 2022.

\bibitem[Wang et~al.(2019)Wang, Yu, Li, Yang, Fu, and Heng]{BEAL}
Shujun Wang, Lequan Yu, Kang Li, Xin Yang, Chi-Wing Fu, and Pheng-Ann Heng.
\newblock Boundary and entropy-driven adversarial learning for fundus image segmentation.
\newblock In \emph{Int. Conf. Med. Image Comput. Comput.-Assist. Intervent.}, pages 102--110. Springer, 2019.

\bibitem[Wang et~al.(2023{\natexlab{a}})Wang, Zhang, Yan, Zhang, and Li]{TSD}
Shuai Wang, Daoan Zhang, Zipei Yan, Jianguo Zhang, and Rui Li.
\newblock Feature alignment and uniformity for test time adaptation.
\newblock In \emph{IEEE Conf. Comput. Vis. Pattern Recog.}, pages 20050--20060, 2023{\natexlab{a}}.

\bibitem[Wang et~al.(2023{\natexlab{b}})Wang, Zhong, Wang, Chen, Ling, Wang, and Sebe]{Dynamically}
Wei Wang, Zhun Zhong, Weijie Wang, Xi Chen, Charles Ling, Boyu Wang, and Nicu Sebe.
\newblock Dynamically instance-guided adaptation: A backward-free approach for test-time domain adaptive semantic segmentation.
\newblock In \emph{IEEE Conf. Comput. Vis. Pattern Recog.}, pages 24090--24099, 2023{\natexlab{b}}.

\bibitem[Yang et~al.(2022)Yang, Chen, Jiang, Liu, Cao, Heng, and Dou]{DLTTA}
Hongzheng Yang, Cheng Chen, Meirui Jiang, Quande Liu, Jianfeng Cao, Pheng~Ann Heng, and Qi Dou.
\newblock Dltta: Dynamic learning rate for test-time adaptation on cross-domain medical images.
\newblock \emph{IEEE Trans. Med. Imaging}, 41\penalty0 (12):\penalty0 3575--3586, 2022.

\bibitem[Yang and Soatto(2020)]{FDA}
Yanchao Yang and Stefano Soatto.
\newblock {FDA}: Fourier domain adaptation for semantic segmentation.
\newblock In \emph{IEEE Conf. Comput. Vis. Pattern Recog.}, pages 4085--4095, 2020.

\bibitem[Zhang et~al.(2023)Zhang, Qi, Shi, and Gao]{DomainAdaptor}
Jian Zhang, Lei Qi, Yinghuan Shi, and Yang Gao.
\newblock Domainadaptor: A novel approach to test-time adaptation.
\newblock In \emph{IEEE Conf. Comput. Vis. Pattern Recog.}, pages 18971--18981, 2023.

\bibitem[Zhang et~al.(2022)Zhang, Levine, and Finn]{MEMO}
Marvin Zhang, Sergey Levine, and Chelsea Finn.
\newblock Memo: Test time robustness via adaptation and augmentation.
\newblock \emph{Adv. Neural Inform. Process. Syst.}, 35:\penalty0 38629--38642, 2022.

\bibitem[Zhang et~al.(2024)Zhang, Zhou, Tao, Wang, Wu, Liu, Gu, Chen, and Chen]{TestFit}
Yizhe Zhang, Tao Zhou, Yuhui Tao, Shuo Wang, Ye Wu, Benyuan Liu, Pengfei Gu, Qiang Chen, and Danny~Z Chen.
\newblock {TestFit}: A plug-and-play one-pass test time method for medical image segmentation.
\newblock \emph{Medical Image Analysis}, 92:\penalty0 103069, 2024.

\bibitem[Zhang et~al.(2010)Zhang, Yin, Liu, Wong, Tan, Lee, Cheng, and Wong]{ORIGA}
Zhuo Zhang, Feng~Shou Yin, Jiang Liu, Wing~Kee Wong, Ngan~Meng Tan, Beng~Hai Lee, Jun Cheng, and Tien~Yin Wong.
\newblock {ORIGA-light}: An online retinal fundus image database for glaucoma analysis and research.
\newblock In \emph{Int. Conf. IEEE Eng. in Med. and Bio.}, pages 3065--3068. IEEE, 2010.

\bibitem[Zhou et~al.(2022)Zhou, Yang, Loy, and Liu]{CoCoOp}
Kaiyang Zhou, Jingkang Yang, Chen~Change Loy, and Ziwei Liu.
\newblock Conditional prompt learning for vision-language models.
\newblock In \emph{IEEE Conf. Comput. Vis. Pattern Recog.}, pages 16816--16825, 2022.

\bibitem[Zhu et~al.(2021)Zhu, Ni, Xu, Kong, Huang, and Goldstein]{Initialize1}
Chen Zhu, Renkun Ni, Zheng Xu, Kezhi Kong, W~Ronny Huang, and Tom Goldstein.
\newblock Gradinit: Learning to initialize neural networks for stable and efficient training.
\newblock \emph{Adv. Neural Inform. Process. Syst.}, 34:\penalty0 16410--16422, 2021.

\bibitem[Zuo et~al.(2021)Zuo, Yao, and Xu]{MSDA}
Yukun Zuo, Hantao Yao, and Changsheng Xu.
\newblock Attention-based multi-source domain adaptation.
\newblock \emph{IEEE Trans. Image Process.}, 30:\penalty0 3793--3803, 2021.

\end{thebibliography}
